\documentclass[10pt,twocolumn,letterpaper]{article}
\ifdefined\pdfminorversion
\fi
\usepackage[hmargin=0.8in,top=0.9in,bottom=1in,columnsep=0.3in]{geometry}
\usepackage{times}
\usepackage{xspace}
\usepackage{xcolor}
\usepackage{graphicx}
\usepackage{amsmath,amssymb}
\usepackage{booktabs}
\usepackage[numbers,sort&compress]{natbib}
\usepackage[format=plain,labelsep=period]{caption}
\usepackage{colortbl}
\usepackage{tabularx}
\usepackage{float}
\usepackage{algorithm}
\usepackage{algpseudocode}
\usepackage{stfloats}
\usepackage{placeins}
\usepackage[hyphens]{url}
\usepackage[breaklinks,colorlinks]{hyperref}

\newcommand{\method}{LoopZip\xspace}

\definecolor{softblue}{HTML}{CFE3F6}
\definecolor{softcyan}{HTML}{CDEFF1}
\definecolor{softgray}{HTML}{F2F2F2}
\definecolor{gainblue}{HTML}{2878B5}

\title{LoopZip: Visual Token Compression for Efficient Looped Vision-Language Models}
\author{Weizhong Huang\textsuperscript{1}\quad Zhicheng Liu\textsuperscript{1}\quad Yan Zhang\textsuperscript{1}\quad Fei Chao\textsuperscript{1}\quad Xiawu Zheng\textsuperscript{1}\quad Liujuan Cao\textsuperscript{1,}\thanks{Corresponding author.}\\
\textsuperscript{1}Key Laboratory of Multimedia Trusted Perception and Efficient Computing,\\
Ministry of Education of China, Xiamen University, 361005, P.R. China.}
\date{}

\newcommand{\mainpaperfloats}{
  \raggedbottom
  \setcounter{dbltopnumber}{1}
  \setcounter{dblbotnumber}{1}
  \renewcommand{\bottomfraction}{0.48}
  \setcounter{topnumber}{1}
  \setcounter{totalnumber}{2}
  \renewcommand{\dbltopfraction}{0.75}
  \renewcommand{\topfraction}{0.48}
  \renewcommand{\textfraction}{0.20}
}

\hypersetup{pdftitle={LoopZip: Visual Token Compression for Efficient Looped Vision-Language Models},pdfauthor={Weizhong Huang, Zhicheng Liu, Yan Zhang, Fei Chao, Xiawu Zheng, Liujuan Cao}}
\begin{document}
\maketitle
\mainpaperfloats

\begin{abstract}
Looped vision--language models (VLMs) deepen computation by reusing shared layers, approaching larger models' capabilities with fewer parameters and easing the storage and deployment burden of stacking independent layers. However, repeatedly processing long visual sequences still incurs substantial compute and GPU memory costs. Visual token compression reduces these costs by shortening sequences. Existing methods mainly select original tokens directly or combine selection with aggregation. In looped models, the former may overlook changing token importance across calls and discard information needed later; the latter lack joint consideration of output changes across calls, limiting information preservation. We propose \method, a training-free visual token compression algorithm addressing both issues. Visual aggregation builds a compact candidate pool before the looped backbone, and cross-invocation selection averages instruction attention from the first two existing calls of the same shared attention layer to retain tokens needed by both states. Weight-tied value correction then jointly fits shared visual-value adjustments to both pre-selection layer outputs, reducing output deviations across states. To validate \method's effectiveness, we conduct extensive experiments covering two open-source looped VLMs, eleven benchmarks, and three pruning ratios. On LoopVL-1B, \method retains \textup{\textbf{97.16\%}} of the original model's performance at $50\%$ pruning ratio and exceeds the strongest baseline by \textup{\textbf{5.36\%}} at $87.5\%$ pruning ratio. Across Hugging Face Transformers, vLLM, and SGLang, LoopVL-1B with \method reduces peak allocated GPU memory by \textup{\textbf{58.36\%--58.98\%}} and achieves \textup{\textbf{2.01--2.47}}$\times$ end-to-end speedups over the original model.
\end{abstract}

\section{Introduction}
\label{sec:intro}

Vision--language models (VLMs) excel in image understanding, visual question answering, and multimodal reasoning~\cite{alayrac2022flamingo,li2023blip2,liu2023llava,chen2024internvl}, with progress partly driven by greater parameter counts and network depth~\cite{liu2024llava15,bai2025qwen25vl}.
However, stacking independent layers increases parameter storage and deployment costs~\cite{huang2025losa,huang2025atp,huang2025shapleymoe,zhang2026spt,huang2026commitkv}.
Looped VLMs repeatedly execute shared layers and update visual and language representations, approaching larger models' performance with fewer parameters~\cite{saunshi2025reasoning,bae2025relaxed,qian2026loopvl,sensenova2026loop}.
Yet long visual sequences from high-resolution images still require repeated processing, incurring high computation and GPU memory costs~\cite{guo2024llavauhd,li2024monkey}.

\begin{figure}[!t]
  \centering
  \includegraphics[width=\columnwidth]{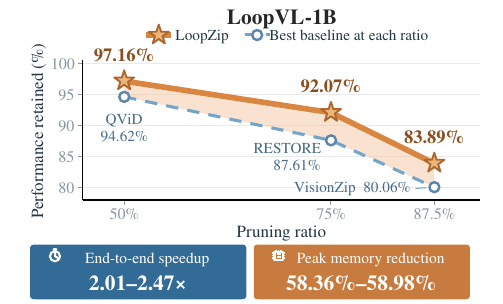}
  \caption{\textbf{Performance and efficiency on LoopVL-1B.} \method maintains high performance under pruning while reducing inference latency and GPU memory usage.}
  \label{fig:first_page_teaser}
\end{figure}

Visual token compression shortens sequences by selecting or aggregating visual information, reducing costs while preserving information needed to answer questions.
Existing methods mainly select original tokens directly or combine selection with aggregation.

\emph{Direct token selection.}
FastV~\cite{chen2024fastv} uses attention, CDPruner~\cite{zhang2025cdpruner} considers instruction-conditioned diversity, and V$^2$Drop~\cite{chen2026v2drop} uses cross-layer hidden-state changes.
IVC-Prune~\cite{sun2026ivcprune} retains coordinate-bearing and semantically relevant tokens, while QViD~\cite{athi2026vision} guides selection through query--vision interaction decomposition.
However, these methods do not explicitly account for cross-invocation changes in visual-token importance and may discard information needed by later calls.

\emph{Selection with aggregation.}
VisionZip~\cite{yang2025visionzip} selects tokens by encoder attention and merges similar remaining tokens into summaries.
VisionTrim~\cite{yu2026visiontrim} uses global importance and local spatial relationships for selection, then merges remaining tokens around text-relevant centers.
SparseVLM~\cite{zhang2025sparsevlm} selects by text attention and recycles more important pruned tokens into new tokens through grouping and merging.
RESTORE~\cite{cho2026restore} selects representative, distinct centers and merges other tokens into their most similar centers, preserving original positions and calibrating attention.
However, these methods do not jointly account for output changes across calls of a shared layer; their aggregated representations may not reduce information loss in looped VLMs.

These limitations motivate compression that considers both changing visual needs and output deviations across calls.
We introduce \method, a training-free algorithm that combines early aggregation with cross-invocation selection and value correction.
\method first builds a smaller candidate pool through visual aggregation before the looped backbone, then selects tokens using mean instruction attention from the first two existing calls of the same shared attention layer, reflecting both states' information needs.
Using both calls' pre-selection layer outputs as targets, it jointly fits one shared visual-value correction to reduce output deviations.
The correction is applied only to retained visual values at the second call; after the current block finishes, the visual sequence is shortened so that later calls benefit from both operations.

We evaluate \method on LoopVL-1B~\cite{qian2026loopvl} and SenseNova-Loop-9B-Preview~\cite{sensenova2026loop}, two open-source looped VLMs, across eleven benchmarks covering visual understanding, hallucination evaluation, and multimodal reasoning, using average scores and relative performance retention.
\method achieves the highest average scores among compared compression methods: on LoopVL-1B, it exceeds the strongest baseline, VisionZip, by $5.36\%$ in average score at $87.5\%$ average pruning and retains $97.16\%$ of the original model's performance at $50\%$ average pruning.
When deployed on Hugging Face Transformers, vLLM, and SGLang, it reduces peak allocated GPU memory by $58.36\%$--$58.98\%$ and achieves $2.01$--$2.47\times$ end-to-end speedups.
These results demonstrate \method's effectiveness and superiority.

\section{Related Work}
\label{sec:related_work}

\paragraph{Looped Models.}
Looped models use shared layers to increase effective depth~\cite{dehghani2019universal,giannou2023looped,csordas2024moeut,graves2016act,lan2020albert}, supporting language and latent reasoning~\cite{geiping2025recurrent,zhu2025ouro,hao2024coconut} and hierarchical or recursive reasoning~\cite{wang2025hrm,jolicoeur2025trm}.
LoopVL~\cite{qian2026loopvl} and SenseNova-Loop~\cite{sensenova2026loop} extend this design to VLMs, but repeated visual-token processing still incurs computation and GPU memory costs; \method reduces these costs by compressing the visual sequence.

\paragraph{Visual Token Compression.}
Existing methods mainly use direct selection or selection with aggregation.
Direct selection uses attention (FastV~\cite{chen2024fastv}), instruction-conditioned diversity (CDPruner~\cite{zhang2025cdpruner}), cross-layer hidden-state changes (V$^2$Drop~\cite{chen2026v2drop}), coordinate and semantic cues (IVC-Prune~\cite{sun2026ivcprune}), or query--vision interactions (QViD~\cite{athi2026vision}).
VisionZip~\cite{yang2025visionzip}, VisionTrim~\cite{yu2026visiontrim}, SparseVLM~\cite{zhang2025sparsevlm}, and RESTORE~\cite{cho2026restore} combine selection with merging or recycling to preserve information from tokens not retained individually.
These methods do not jointly account for changing visual needs and output deviations across calls of the same shared layer.
\method combines visual aggregation, cross-invocation selection, and weight-tied value correction.

\begin{figure*}[t]
  \centering
  \includegraphics[width=\textwidth]{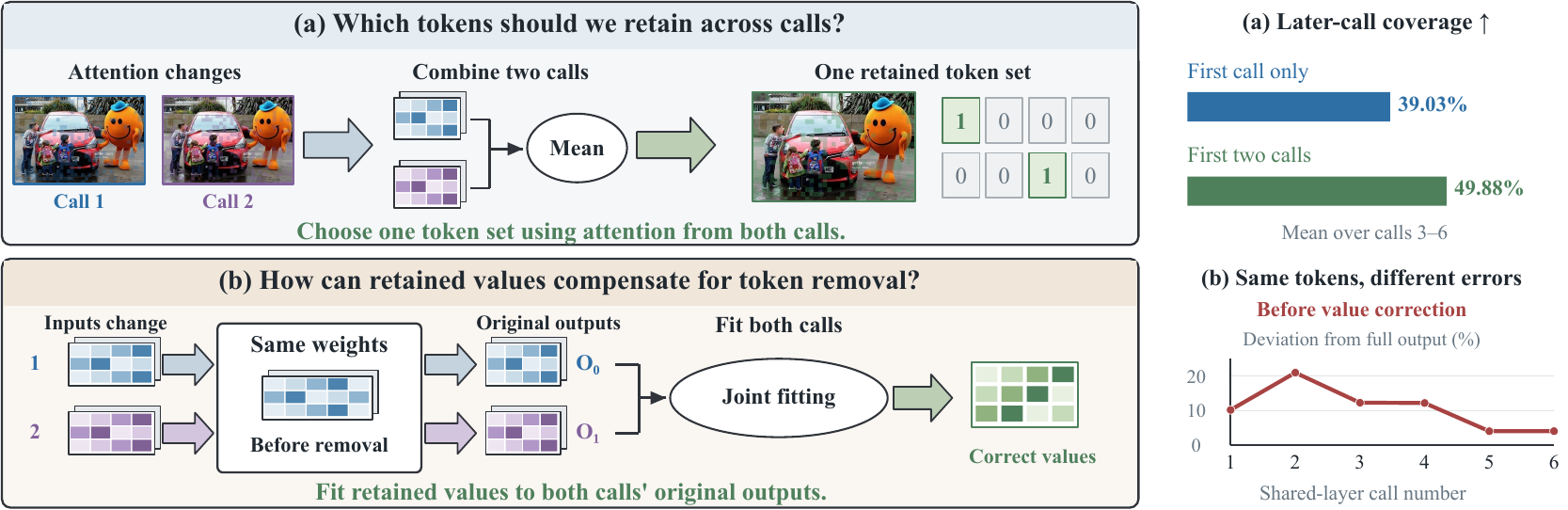}
  \caption{\textbf{Why should token compression consider different calls?} (a) Combining two calls improves later-token coverage. (b) Right: removal-induced output deviation for one fixed first-call selection, before value correction; the horizontal axis counts calls of the same shared layer. Left: the motivated joint correction. Diagnostics use LoopVL-1B, $25\%$ retention, and means over $160$ questions; Section~\ref{sec:further_analysis} gives the full analysis.}
  \label{fig:loopzip_motivation}
\end{figure*}

\begin{figure*}[t]
  \centering
  \includegraphics[width=\textwidth]{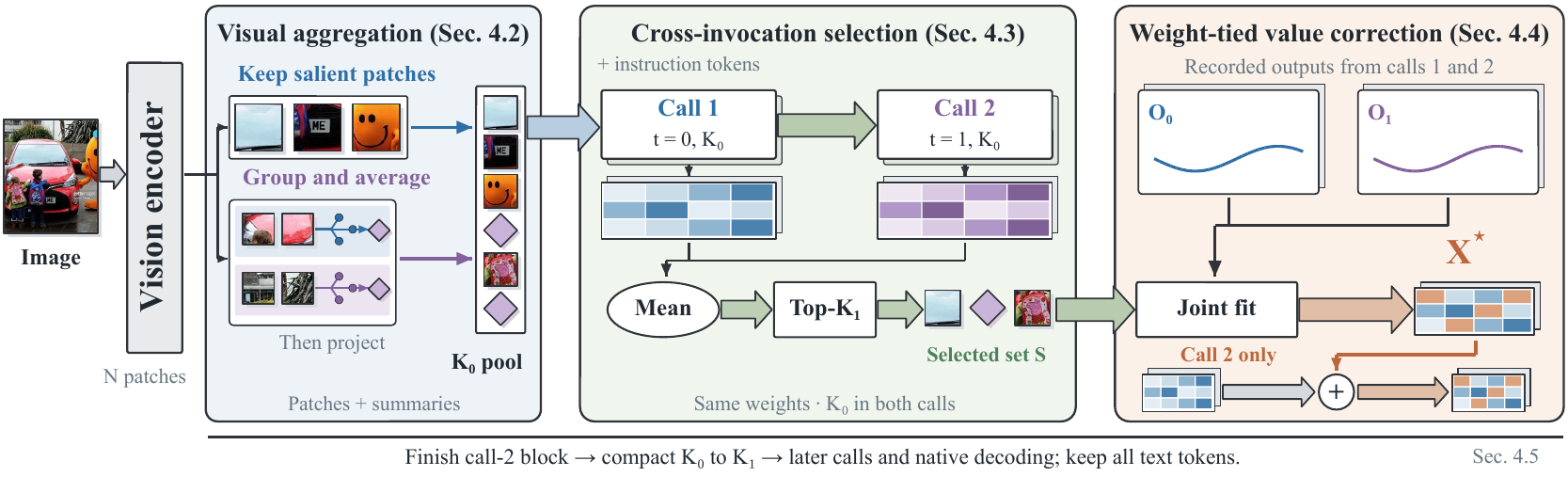}
  \caption{\textbf{\method's forward path.} Aggregation forms $K_0$ candidates from $N$ patches (Sec.~\ref{sec:recurrent_compression}). Both first calls receive $K_0$ candidates; their mean attention selects $K_1$ (Sec.~\ref{sec:response_contribution}). Their outputs $\boldsymbol O_0,\boldsymbol O_1$ jointly supervise correction $\boldsymbol X^\star$, applied only to retained values at call $2$ (Sec.~\ref{sec:value_correction}). After that block, later calls use $K_1$ visual tokens (Sec.~\ref{sec:loopzip_algorithm}); all text tokens remain.}
  \label{fig:loopzip_overview}
\end{figure*}

\section{Preliminaries}
\label{sec:preliminaries}

\subsection{Visual Token Pruning in Looped VLMs}
\label{sec:recurrent_vlm}

A looped VLM comprises a vision encoder, a multimodal projector, and a recurrent language backbone~\cite{qian2026loopvl,sensenova2026loop}.
The encoder extracts $N$ image patch features $\boldsymbol{f}_i$, which the pretrained projector $\mathcal P$ maps to visual embeddings $\boldsymbol{e}_i=\mathcal P(\boldsymbol{f}_i)\in\mathbb R^{d_{\mathrm{model}}}$, where $d_{\mathrm{model}}$ is the backbone's hidden dimension.
Tokenized instructions are embedded through the text embedding table and combined with the visual embeddings to form the input sequence.

The backbone repeatedly invokes shared Transformer modules~\cite{vaswani2017attention} following a fixed schedule.
Let $\boldsymbol{s}^{(0)}$ be the initial model state formed from the visual and instruction embeddings. Subsequent updates are
\begin{equation}
  \boldsymbol{s}^{(r)}=\mathcal F_{\boldsymbol{\theta},r}\!\left(\boldsymbol{s}^{(r-1)}\right),
  \qquad r=1,\ldots,R.
  \label{eq:method_recurrence}
\end{equation}
Here $\mathcal F_{\boldsymbol{\theta},r}$ is the module scheduled at step $r$, $R$ is the total number of module calls, and $\boldsymbol{\theta}$ denotes the fixed pretrained parameters.
Each module contains several Transformer layers; repeated calls reuse the same layer weights but process the updated model state.
Thus, the same shared layer receives different input states across calls.

Visual token compression reduces $N$ visual tokens to $K<N$ through selection or aggregation.
Aggregation averages similar patch features before projection to form representative visual embeddings~\cite{yang2025visionzip}.
The visual token budget is $K$, and the pruning ratio is $p=1-K/N$.
Compression preserves all instruction tokens and the original loop schedule, but using fewer visual tokens reduces subsequent computation and GPU memory consumption~\cite{yang2026flashloop}.

\subsection{Motivation}
\label{sec:observations}

Shared layers process changing inputs across calls (Section~\ref{sec:recurrent_vlm}). \textbf{What does this reuse require of token compression?} We examine selection and value correction on LoopVL-1B's first shared attention layer (Figure~\ref{fig:loopzip_motivation}).

\paragraph{Which tokens should we retain across calls?}
Single-call attention may miss other calls' needs; multiple calls can accommodate them. We compare first-call attention with the first two calls' mean attention at the same budget. Figure~\ref{fig:loopzip_motivation}a shows that two calls better cover each later call's high-attention tokens. Thus, \textbf{we combine attention across calls to score tokens and select one retained set.}

\paragraph{How can retained values compensate for token removal?}
Identical token removal may affect calls differently. We test this with a fixed $25\%$ visual-token subset selected by first-call attention. On each recorded original state, we remove the rest without value correction and compare with that call's uncompressed output. Varying deviations (Figure~\ref{fig:loopzip_motivation}b, right) show that one call's output cannot capture effects on others. Thus, \textbf{we use original outputs across calls to jointly correct retained token values.}

\paragraph{How many calls should inform token scoring and value correction?}
Both steps need cross-call information: one call cannot reveal changes between calls; two are the minimum. More observations add states but delay scoring, correction, and compaction, postponing benefits to later computation. Table~\ref{tab:ablation_observation_count} reports $55.47$--$95.67$\,ms for three to six observations versus $42.05$\,ms for two; accuracy does not improve monotonically among newly evaluated settings. Thus, \textbf{we use the first two calls for token scoring and value correction, capturing cross-call information while completing compression early.}

\section{Methodology}
\label{sec:method}

\subsection{Insight and Method Overview}
\label{sec:method_overview}

Figure~\ref{fig:loopzip_overview} implements Section~\ref{sec:observations}'s design: \textbf{select tokens using both calls' attention} (Section~\ref{sec:response_contribution}), then \textbf{jointly correct their values using both original outputs} (Section~\ref{sec:value_correction}). The first two existing calls allow early selection and correction.

Before the backbone, visual aggregation lowers early-call cost while preserving candidates for instruction-guided selection (Section~\ref{sec:recurrent_compression}). Token counts follow:
\begin{equation}
 N\xrightarrow{\text{aggregation}}K_0\xrightarrow{\text{selection}}K_1,
 \quad K_1<K_0<N.
 \label{eq:method_token_stages}
\end{equation}
$N$, $K_0$, and $K_1$ are the original patch, candidate, and final counts, respectively.
Both observed calls receive $K_0$ tokens.
At call $2$, we select $K_1$ tokens and correct only their visual values.
After the current block, we compact to $K_1$ for later calls and native cached decoding (Section~\ref{sec:loopzip_algorithm}).
All text tokens, pretrained weights, and the loop schedule are preserved (Algorithm~\ref{alg:loopzip}).

\subsection{Visual Aggregation}
\label{sec:recurrent_compression}
\begin{figure}[!htbp]
  \centering
  \includegraphics[width=\columnwidth]{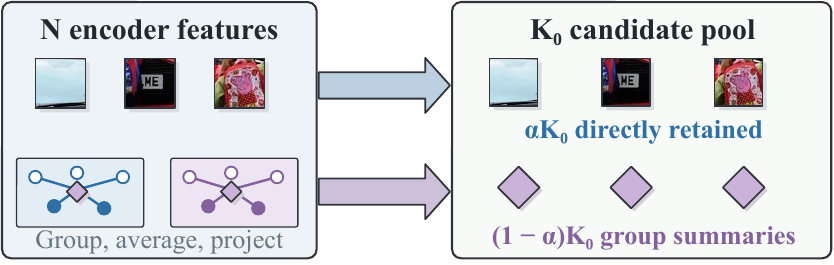}
  \caption{\textbf{Visual aggregation (schematic).} High-attention patches are retained directly (image tiles); similar remaining features form projected group summaries (diamonds). Both types fill the $K_0$-token candidate pool.}
  \label{fig:method_visual_aggregation}
\end{figure}

To reduce computation before cross-invocation selection, we follow VisionZip~\cite{yang2025visionzip} and aggregate the $N$ original patch features into $K_0<N$ candidate visual tokens.
The pool contains directly retained tokens and one summary token per group, so $K_0$ equals the number of directly retained patches plus the number of groups.
We allocate a fraction $\alpha$ of the $K_0$ candidate slots to directly retain patches with high encoder attention; the remaining slots are used for group summaries.
Varying $\alpha$ changes only the mix of the two token types, not the total candidate count.

We set $K_0$ and $K_1$ jointly to meet the target average pruning budget.
Let $L$ be the total number of layer calls in the language backbone, $d$ the number charged with $K_0$ visual tokens, and $p$ the target average pruning ratio. Then
\begin{equation}
 \frac{dK_0+(L-d)K_1}{L}\approx(1-p)N.
 \label{eq:method_average_budget}
\end{equation}
The left side is the average visual token count over both stages; the right side is the target average retained count.
The budget includes both early calls using $K_0$ and later calls using $K_1$; the approximation comes from integer rounding.

We first retain high-scoring patches according to attention scores in the vision encoder's final layer, keeping their original embeddings $\boldsymbol{e}_i$.
The number of remaining pool slots determines the number of centers, sampled at a fixed stride from the remaining patches in image order.
Other patches are assigned to their most similar center by cosine similarity of encoder output features.

Let $c$ index a center and $\mathcal C_c$ contain that center and its assigned patches.
Each group produces one summary token:
\begin{equation}
  \widetilde{\boldsymbol{e}}_c=\mathcal P\!\left(
       \frac{1}{|\mathcal C_c|}
       \sum_{i\in\mathcal C_c}\boldsymbol{f}_i\right).
 \label{eq:method_static_summary}
\end{equation}
Here $|\mathcal C_c|$ is the number of patches in the group, and the summary embedding $\widetilde{\boldsymbol{e}}_c$ inherits the center's original spatial position.
The retained embeddings and group summaries are ordered by their original positions to form the $K_0$-token pool, from which the next stage selects $K_1$ tokens (Figure~\ref{fig:method_visual_aggregation}).

\subsection{Cross-Invocation Selection}
\label{sec:response_contribution}
\begin{figure}[!htbp]
  \centering
  \includegraphics[width=\columnwidth]{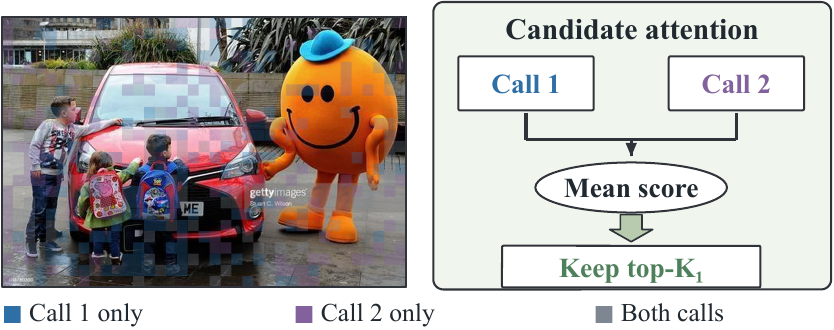}
  \caption{\textbf{Cross-invocation selection.} The car image overlays first-call-only (blue), second-call-only (purple), and shared (gray) top-attention patches. The schematic combines the two calls\textquotesingle{} candidate attention scores and retains $K_1$ tokens.}
  \label{fig:method_attention_dynamics}
\end{figure}

The aggregated $K_0$ candidates and instruction tokens enter the looped backbone.
The evolving state $\boldsymbol{s}^{(r)}$ changes the shared layer's inputs and attention, so we compare candidates across calls before making the final selection (Figure~\ref{fig:method_attention_dynamics}).
We select one shared attention layer within the looped module and combine attention from its first two existing calls for token selection.
For LoopVL-1B, we use the first attention layer in the shared low-level module, taking the first two of its six calls; for SenseNova-Loop-9B-Preview, we use the first full-attention layer inside the looped block, taking its only two calls.
We average attention from these calls to select $K_1$ visual tokens from the candidate pool.
Both observations come from the original loop schedule and require no extra module executions.

Directly retained and summary tokens are scored together.
Index candidate positions by $i=1,\ldots,K_0$ and the two observed calls by $t=0,1$.
At both calls, we use the same $P$ instruction positions and average attention over these positions and heads to obtain $\bar a_{t,i}$ for candidate $i$.
We retain the native attention normalization; if the instruction is empty, we use nonvisual prefix positions.
We average the two scores and select the top $K_1$ candidates:
\begin{equation}
 s_i=\frac{\bar a_{0,i}+\bar a_{1,i}}{2},\qquad
 \mathcal S=\operatorname{TopK}_{i}(s_i;K_1).
 \label{eq:method_selection_score}
\end{equation}
Here $s_i$ is the combined score, $\mathcal S$ is the retained index set, and $K_1<K_0$.
Both candidate types use the same selection rule, so initially retained tokens may also be removed.
Ties favor earlier pool positions; the original order and all text tokens are preserved.
The next stage corrects the selected visual values to reduce output changes caused by removing the other tokens.

\subsection{Weight-Tied Value Correction}
\label{sec:value_correction}
\begin{figure}[!htbp]
  \centering
  \includegraphics[width=\columnwidth]{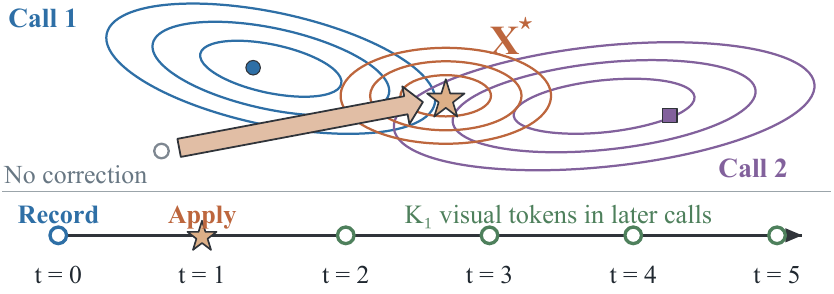}
  \caption{\textbf{Shared value correction (schematic).} The two calls' fitting terms (blue/purple) yield one regularized joint solution $\mathbf X^\star$ (orange). The LoopVL timeline shows that only call $t=1$ receives the correction.}
  \label{fig:method_joint_correction}
\end{figure}

Cross-invocation selection determines which $K_1$ candidates to retain, but removing the others still changes attention outputs.
To address this second issue, we adjust the retained tokens' value vectors using both observed states, keeping their compressed outputs close to the corresponding full-pool outputs.

We use the same two calls $t=0,1$ and the same $P$ query positions as in the preceding section.
Let $\mathbf O_t$ be the full-pool output at these positions, and let $\widehat{\mathbf O}_t(\mathbf X)$ be the estimated output after retaining the selected visual tokens and all text tokens and applying the value correction $\mathbf X$.
The two calls share the same layer parameters, so we jointly solve for one correction matrix:
\begin{equation}
 \mathbf X^\star=\operatorname*{arg\,min}_{\mathbf X}\;
   \frac{1}{2P}\sum_{t=0}^{1}
    \|\widehat{\mathbf O}_t(\mathbf X)-\mathbf O_t\|_F^2
   +\lambda\|\mathbf X\|_F^2.
 \label{eq:method_response_fit}
\end{equation}
The first term keeps each call's output close to its original output, while the second limits the correction magnitude.
We solve this problem using conjugate gradients~\cite{hestenes1952conjugate}.

The value correction is jointly fitted from the shared layer's first two calls and is solved only after the second call ($t=1$) produces its output.
By then, the first call's ($t=0$) forward computation has finished; correcting its forward output would require rerunning it, adding computation.
We therefore keep the first call's output and cache unchanged and correct only the retained visual values at the second call.
Figure~\ref{fig:method_joint_correction} illustrates the shared fit and its application timing.

\begin{algorithm}[!t]
  \caption{LoopZip compression procedure}
  \label{alg:loopzip}
  \normalsize
  \algrenewcommand\algorithmicindent{0.8em}
  \begin{algorithmic}[1]
    \Require Image, instruction, frozen model, target average pruning ratio
    \Ensure Answer using compressed visual tokens
    \State Build a $K_0$-token pool by visual aggregation (Sec.~\ref{sec:recurrent_compression})
    \State Keep all text tokens
    \For{$t=0,\ldots,T-1$}
      \State Advance the original schedule to call $t$ of the shared layer, using the current sequence (Sec.~\ref{sec:recurrent_vlm})
      \If{$t<2$}
        \State Record instruction attention and the $K_0$-pool output (Secs.~\ref{sec:response_contribution}--\ref{sec:value_correction})
      \EndIf
      \If{$t=1$}
        \State Average both calls' attention and select $K_1$ visual tokens (Sec.~\ref{sec:response_contribution})
        \State Jointly fit one value correction to both recorded outputs (Sec.~\ref{sec:value_correction})
        \State Apply the correction to the retained visual values at this call (Sec.~\ref{sec:value_correction})
      \EndIf
      \State Finish the current block (Sec.~\ref{sec:loopzip_algorithm})
      \If{$t=1$}
        \State Shorten the visual sequence to the selected $K_1$ tokens, keeping all text tokens (Sec.~\ref{sec:loopzip_algorithm})
      \EndIf
    \EndFor
    \State Finish the remaining layer calls and use native cached decoding to generate the answer (Sec.~\ref{sec:loopzip_algorithm})
  \end{algorithmic}
  \par\smallskip
  {\footnotesize\noindent\textit{Note.} $T$ is the selected shared attention layer's total call count in one forward pass: $T=6$ for LoopVL-1B and $T=2$ for SenseNova-Loop-9B-Preview. Calls are indexed from $t=0$; only the first two are used for token selection and value fitting.\par}
\end{algorithm}

\subsection{Execution with Compressed Tokens}
\label{sec:loopzip_algorithm}

Following the value fitting in Section~\ref{sec:value_correction}, we apply the correction at the shared layer's second call ($t=1$).
Once the current block finishes, we shorten the visual sequence from $K_0$ tokens to the selected $K_1$, keeping all text tokens.
Subsequent calls use the shorter sequence with the original loop schedule and cached decoding, reducing computation and GPU memory use.
Algorithm~\ref{alg:loopzip} summarizes the full procedure.

\section{Experiments}
\label{sec:experiments}

\begin{table*}[t]
 \centering
 \caption{\textbf{Two looped VLMs at $50\%$ pruning ratio.} Scores are percentages; Rel.\ Avg.\ averages ratios to Vanilla. Bold and underlined values indicate the best and second-best results among compression methods, respectively.}
 \label{tab:models}
 \setlength{\tabcolsep}{1pt}
 \renewcommand{\arraystretch}{0.88}
 \footnotesize
 \begin{tabularx}{\textwidth}{@{}l*{13}{>{\centering\arraybackslash}X}@{}}
\toprule
 & \multicolumn{6}{c}{\textbf{Visual Understanding}} & \multicolumn{2}{c}{\textbf{Hallucination}} & \multicolumn{3}{c}{\textbf{Reasoning}} & \multicolumn{2}{c}{} \\
\cmidrule(lr){2-7}\cmidrule(lr){8-9}\cmidrule(lr){10-12}
\textbf{Method} & \textbf{\shortstack{MM\\Star}} & \textbf{RWQA} & \textbf{VMCB} & \textbf{AI2D} & \textbf{\shortstack{Chart\\QA}} & \textbf{\shortstack{HR\\Bench\\4K}} & \textbf{POPE} & \textbf{Hallu.} & \textbf{\shortstack{MMMU\\-Pro}} & \textbf{MathV} & \textbf{LogicV} & \textbf{Avg.$\uparrow$} & \textbf{\shortstack{Rel.\\Avg.$\uparrow$}} \\
\midrule
\multicolumn{14}{c}{\cellcolor{softblue!45}\textbf{LoopVL-1B~\cite{qian2026loopvl}}} \\
\midrule
\rowcolor{softgray}Vanilla & 63.13 & 70.85 & 70.50 & 75.78 & 74.60 & 56.50 & 90.84 & 44.41 & 38.32 & 38.22 & 43.18 & 60.58 & 100.00\% \\
\midrule
FastV~\cite{chen2024fastv}~(ECCV'24) & 53.00 & 59.87 & 66.00 & 69.95 & 45.16 & 51.25 & 85.80 & 38.65 & 35.03 & 34.51 & 38.26 & 52.50 & 87.04\% \\
SparseVLM~\cite{zhang2025sparsevlm}~(ICML'25) & 54.60 & 58.69 & 65.30 & 70.01 & 48.48 & 51.62 & 85.69 & 38.45 & 35.90 & 34.41 & 37.81 & 52.81 & 87.53\% \\
CDPruner~\cite{zhang2025cdpruner}~(NeurIPS'25) & 56.07 & 67.06 & \underline{68.20} & 73.41 & 49.04 & 52.62 & 89.56 & 38.86 & 35.49 & 32.43 & 37.14 & 54.53 & 89.59\% \\
VisionZip~\cite{yang2025visionzip}~(CVPR'25) & \underline{59.40} & \underline{68.37} & \underline{68.20} & \underline{73.51} & 60.20 & 57.00 & \underline{90.78} & 43.54 & \textbf{37.17} & 36.15 & 36.02 & 57.30 & 94.44\% \\
VisionTrim~\cite{yu2026visiontrim}~(ICLR'26) & 56.67 & 64.58 & 65.90 & 72.99 & 55.60 & 54.75 & 89.30 & 40.77 & 36.53 & 34.01 & 36.47 & 55.23 & 91.00\% \\
IVC-Prune~\cite{sun2026ivcprune}~(ICLR'26) & 56.73 & 61.44 & 64.20 & 70.60 & 47.52 & 54.25 & 88.86 & 38.12 & 35.49 & 34.47 & 36.47 & 53.47 & 88.31\% \\
V$^2$Drop~\cite{chen2026v2drop}~(CVPR'26) & 53.73 & 65.88 & 67.50 & 71.41 & 56.52 & 55.62 & 88.74 & 41.45 & 33.47 & 32.89 & \underline{39.37} & 55.14 & 90.72\% \\
RESTORE~\cite{cho2026restore}~(ICML'26) & 58.60 & 68.10 & 66.90 & 73.12 & 65.24 & 56.88 & 90.12 & 41.87 & 36.59 & \textbf{36.68} & 37.58 & \underline{57.43} & 94.58\% \\
QViD~\cite{athi2026vision}~(NeurIPS'26) & 57.67 & 64.58 & 66.80 & 71.28 & \underline{65.72} & \textbf{59.88} & 89.96 & \textbf{44.08} & 35.95 & 35.82 & 38.70 & 57.31 & \underline{94.62\%} \\
\addlinespace[2pt]
\rowcolor{softcyan}\textbf{\method (Ours)} & \textbf{59.73} & \textbf{69.28} & \textbf{70.20} & \textbf{74.77} & \textbf{69.92} & \underline{57.62} & \textbf{90.80} & \underline{43.64} & \underline{36.99} & \underline{36.48} & \textbf{39.82} & \textbf{59.02} & \textbf{97.16\%} \\
\midrule
\multicolumn{14}{c}{\cellcolor{softblue!45}\textbf{SenseNova-Loop-9B-Preview~\cite{sensenova2026loop}}} \\
\midrule
\rowcolor{softgray}Vanilla & 70.60 & 72.42 & 84.50 & 86.24 & 85.16 & 77.00 & 87.56 & 58.28 & 47.98 & 40.79 & 61.74 & 70.21 & 100.00\% \\
\midrule
FastV~\cite{chen2024fastv}~(ECCV'24) & 58.00 & 69.15 & 74.70 & 77.59 & 56.24 & 76.38 & 85.46 & 51.45 & 39.02 & 34.44 & 37.36 & 59.98 & 84.85\% \\
SparseVLM~\cite{zhang2025sparsevlm}~(ICML'25) & 62.07 & 71.24 & 79.70 & 81.02 & 74.56 & \underline{78.12} & 87.83 & 56.39 & 41.39 & 35.20 & 41.16 & 64.43 & 90.90\% \\
CDPruner~\cite{zhang2025cdpruner}~(NeurIPS'25) & 59.67 & 64.97 & 78.90 & 80.86 & 70.32 & 76.50 & 87.40 & 52.22 & 41.10 & 34.97 & 42.95 & 62.71 & 88.52\% \\
VisionZip~\cite{yang2025visionzip}~(CVPR'25) & 59.33 & 66.27 & 75.30 & 78.27 & 65.08 & 74.50 & 85.88 & 50.62 & 39.65 & 35.16 & 41.39 & 61.04 & 86.31\% \\
VisionTrim~\cite{yu2026visiontrim}~(ICLR'26) & 58.13 & 67.45 & 78.30 & 79.86 & 71.64 & 75.50 & 86.24 & 53.95 & 40.69 & 35.03 & 44.52 & 62.85 & 88.80\% \\
IVC-Prune~\cite{sun2026ivcprune}~(ICLR'26) & 60.13 & \textbf{71.63} & 79.30 & 79.86 & 63.76 & 75.50 & 87.27 & 52.56 & 42.72 & 34.38 & 42.51 & 62.69 & 88.68\% \\
V$^2$Drop~\cite{chen2026v2drop}~(CVPR'26) & 62.73 & 70.85 & 80.00 & 81.57 & 79.72 & 76.38 & \textbf{88.11} & 54.80 & 43.06 & \underline{38.09} & 44.07 & 65.40 & 92.54\% \\
RESTORE~\cite{cho2026restore}~(ICML'26) & \underline{65.47} & \underline{71.50} & \underline{80.20} & \underline{81.93} & \underline{80.80} & 77.12 & \underline{87.93} & \underline{57.36} & \underline{44.34} & 37.30 & \underline{46.09} & \underline{66.37} & \underline{93.98\%} \\
QViD~\cite{athi2026vision}~(NeurIPS'26) & 62.13 & 70.33 & 78.30 & 79.44 & 70.52 & 76.25 & 87.49 & 54.15 & 41.68 & 35.56 & 44.07 & 63.63 & 90.00\% \\
\addlinespace[2pt]
\rowcolor{softcyan}\textbf{\method (Ours)} & \textbf{66.33} & \underline{71.50} & \textbf{80.90} & \textbf{83.13} & \textbf{83.28} & \textbf{78.25} & 87.46 & \textbf{58.23} & \textbf{45.61} & \textbf{38.68} & \textbf{50.78} & \textbf{67.65} & \textbf{96.02\%} \\
\bottomrule
\end{tabularx}

\end{table*}

\subsection{Experimental Setup}
\label{sec:experimental_setup}

\begin{figure*}[!b]
  \centering
  \includegraphics[width=\textwidth]{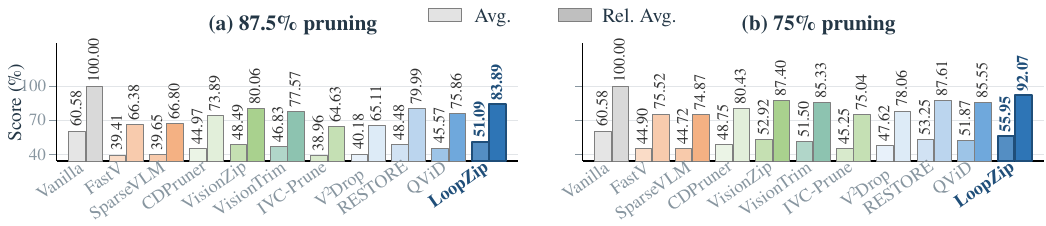}
  \caption{\textbf{Aggregate performance at $87.5\%$ (a) and $75\%$ (b) pruning.} Light/dark bars show Avg.\ and Rel.\ Avg.\ (\%), with values above bars and Vanilla as the unpruned reference. Full results: Table~10, Appendix~E.}
  \label{fig:aggregate_results}
\end{figure*}

\begin{figure*}[!t]
 \centering
 \includegraphics[width=\textwidth]{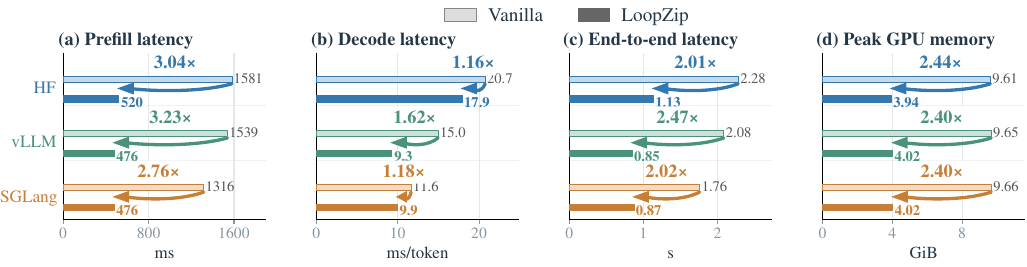}
 \caption{\textbf{Vanilla versus LoopZip at $8192$ visual tokens and $87.5\%$ pruning.} (a)--(d) Prefill, decode, end-to-end latency, and peak allocated GPU memory; lower is better. Colors identify HF, vLLM, and SGLang; light/dark bars show Vanilla/LoopZip. Arrows report Vanilla/LoopZip ratios; labels average five inputs. Full comparisons: Appendix~F.}
 \label{fig:efficiency}
\end{figure*}

\paragraph{Models.}
We evaluate two open-source looped VLMs.
LoopVL-1B~\cite{qian2026loopvl} splits its $32$ backbone layers into $16$-layer low-level and high-level modules. Each forward pass has $2$ cycles, each calling the low module $3$ times and the high module once: $8$ module invocations and $128$ layer calls.
SenseNova-Loop-9B-Preview~\cite{sensenova2026loop} invokes its middle $24$ backbone layers twice with shared weights and the other $8$ once, totaling $32$ layers and $56$ layer calls (App.~D).

\paragraph{Baselines.}
We compare Vanilla and nine compression methods: FastV~\cite{chen2024fastv}, SparseVLM~\cite{zhang2025sparsevlm}, VisionZip~\cite{yang2025visionzip}, CDPruner~\cite{zhang2025cdpruner}, VisionTrim~\cite{yu2026visiontrim}, V$^2$Drop~\cite{chen2026v2drop}, IVC-Prune~\cite{sun2026ivcprune}, RESTORE~\cite{cho2026restore}, and QViD~\cite{athi2026vision}.
Within each model, methods share prompts, image preprocessing, and generation limits at matched pruning budgets (App.~G).

\paragraph{Benchmarks.}
Our eleven benchmarks cover visual understanding (MMStar~\cite{chen2024mmstar}, RealWorldQA~\cite{xai2024realworldqa}, VMCBench~\cite{zhang2025vmcbench}, AI2D~\cite{kembhavi2016ai2d}, ChartQA~\cite{masry2022chartqa}, high-resolution HRBench4K~\cite{wang2025hrbench}), hallucination (POPE~\cite{li2023pope}, HallusionBench~\cite{guan2024hallusionbench}), and reasoning (MMMU-Pro~\cite{yue2025mmmupro}, MathVision~\cite{wang2024mathvision}, LogicVista~\cite{xiao2024logicvista}).
We report mean scores and mean per-benchmark ratios to Vanilla; metrics, dataset sizes, and generation budgets appear in App.~H.

\paragraph{Implementation Details.}
We use NVIDIA H20 96GB GPUs, BF16~\cite{kalamkar2019bfloat16,micikevicius2018mixed}, batch size $1$, greedy decoding, and $50\%$, $75\%$, and $87.5\%$ average pruning.
LoopVL-1B efficiency tests use PyTorch FlashAttention prefill in HF, FlashAttention-2 prefill and native PagedAttention decoding in vLLM, and FlashInfer prefill and paged decoding through SGLang's RadixAttention interface. Serving decoders use native fused operators, a full-token CUDA Graph, and original unquantized weights (integration and output checks: App.~F).
Vanilla and pruned SenseNova use a $1024$-token output ceiling; LoopVL-1B uses input-dependent limits of $8$, $32$, or $512$ tokens (Table~14, App.~H).
Across benchmarks and pruning ratios, aggregation directly retains high-attention patches in $\alpha=0.85$ of candidate slots and uses the rest for group summaries; selection and correction use at most $P=8$ instruction tokens, with correction regularization $\lambda=10^{-4}$.

\subsection{Main Results}
\label{sec:main_results}

\paragraph{Comparison with State-of-the-Art Methods.}
Table~\ref{tab:models} compares methods at $50\%$ average pruning.
On LoopVL-1B, \method leads compression methods with $59.02$ Avg.\ and $97.16\%$ Rel.\ Avg., exceeding the strongest baselines on each metric, RESTORE and QViD, by $1.59$ and $2.54$ percentage points.
It ranks first on $7$ benchmarks and second on the other $4$.
On SenseNova-Loop-9B-Preview, it also leads compression methods with $67.65$ Avg.\ and $96.02\%$ Rel.\ Avg., exceeding RESTORE by $1.28$ and $2.04$ points and ranking first on $9$ of eleven benchmarks.
These results demonstrate \method's effectiveness in preserving looped VLM performance under visual-token compression.

\paragraph{Results at Different Pruning Ratios.}
\method also leads Avg.\ and Rel.\ Avg.\ on LoopVL-1B at $75\%$ and $87.5\%$ average pruning (Figure~\ref{fig:aggregate_results}).
At $75\%$, its $55.95$ Avg.\ and $92.07\%$ Rel.\ Avg.\ exceed RESTORE, the strongest baseline, by $2.70$ and $4.46$ percentage points.
At $87.5\%$, it retains $83.89\%$ of Vanilla performance versus the strongest baseline VisionZip's $80.06\%$, raising Avg.\ from $48.49$ to $51.09$.
This supports robust performance preservation across pruning ratios; per-benchmark results appear in Table~10, App.~E.

\subsection{Efficiency Analyses}
\label{sec:efficiency_analyses}

We compare Vanilla, nine compression baselines, and \method on NVIDIA H20 96GB GPUs under HF, vLLM, and SGLang, greedily generating $32$ tokens on identical inputs.
At $8192$ visual tokens and $87.5\%$ average pruning (Figure~\ref{fig:efficiency}), \method's end-to-end speedups over each framework's Vanilla are $2.01\times$, $2.47\times$, and $2.02\times$, respectively.
Peak allocated memory is $3.94$, $4.02$, and $4.02$\,GiB, down $58.98\%$, $58.41\%$, and $58.36\%$, respectively.
Its vLLM and SGLang speedups over matched same-GPU HF FlashAttention controls are $1.34\times$ and $1.30\times$, respectively.
Prefill and cached decoding accelerate by $2.76$--$3.23\times$ and $1.16$--$1.62\times$, demonstrating reduced inference latency and GPU memory usage.
Tables~11, 12, and~13 in App.~F report all compression methods at $2048$, $4096$, and $8192$ visual tokens and $50\%$, $75\%$, and $87.5\%$ pruning.

\begin{figure}[!htbp]
 \centering
 \includegraphics[width=\columnwidth]{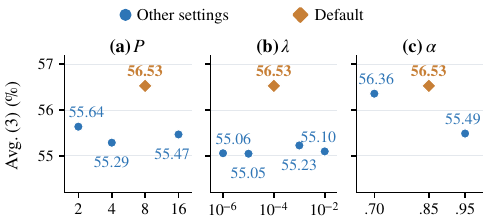}
 \caption{\textbf{Hyperparameter comparison on LoopVL-1B at $87.5\%$ pruning.} Avg.\ (3) averages MMStar, MMMU-Pro, and POPE. Each panel varies one parameter. Blue circles denote other settings; orange diamonds denote the default setting.}
 \label{fig:hyperparameter_main}
\end{figure}

\begin{figure*}[!t]
  \centering
  \includegraphics[width=\textwidth]{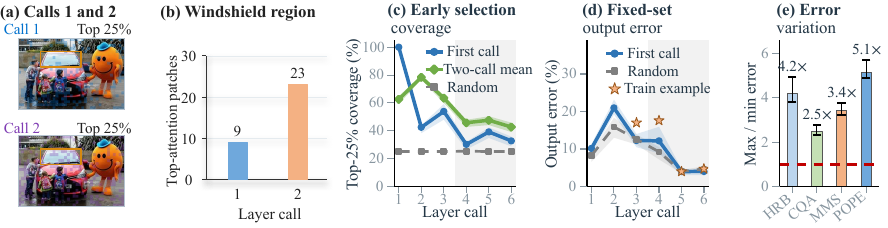}
  \caption{\textbf{LoopVL-1B's first shared attention layer, calls $1$--$6$.}
  (a) First-two top-$25\%$ visual-token masks for an MMStar car example; rectangles mark the same windshield region. (b) Number of top-attention patches within that region.
  At $75\%$ pruning, (c) top-$25\%$ visual-token coverage and (d) layer-output error relative to each call's uncompressed output average $160$ questions. Blue uses first-call attention, green the first two calls' mean attention, and gray random selection. Selected sets stay fixed across calls; stars mark a separate train diagnostic.
  (e) With first-call selection, typical output-error variation is $4.2\times$, $2.5\times$, $3.4\times$, and $5.1\times$ on HRBench4K, ChartQA, MMStar, and POPE; the dashed line denotes equal errors.
  See Appendices~B and~A.}
  \label{fig:observation_detailed}
  \label{fig:further_analysis}
\end{figure*}

\subsection{Ablation Studies}
\label{sec:ablation_studies}
\paragraph{Effectiveness of Components.}
Table~\ref{tab:ablation_static} evaluates components on LoopVL-1B at $87.5\%$ target average pruning, reporting MMStar/MMMU-Pro accuracy, POPE F1, and their average.
Full \method achieves $56.53\%$ Avg.\ (3).
Visual aggregation alone compresses to the target count before the backbone, omitting selection and correction ($55.67\%$).
Removing aggregation instead sends all $N$ visual tokens into the backbone, retaining selection, correction, and the original $K_1$ ($49.76\%$).
Removing cross-invocation selection uses only the second call's attention ($55.45\%$); removing joint fitting fits only that call's output ($55.17\%$).
Removing value correction skips fitting and value updates ($55.28\%$).
All variants score below Full, supporting the overall design's effectiveness in preserving model performance.

\begingroup
\setlength{\intextsep}{1.5pt}
\begin{table}[!htbp]
\centering
\caption{\textbf{Ablation results for individual components (\%).}}
\label{tab:ablation_static}
\setlength{\tabcolsep}{0.5pt}
\setlength{\aboverulesep}{1pt}
\setlength{\belowrulesep}{1pt}
\renewcommand{\arraystretch}{.90}
\small
\begin{tabularx}{\columnwidth}{>{\raggedright\arraybackslash}Xrrrr}
\toprule
Variant & MMStar & \shortstack{MMMU\\-Pro} & POPE & Avg. (3) \\
\midrule
\rowcolor{softcyan}Full \method & \textbf{51.00} & \textbf{31.62} & \textbf{86.98} & \textbf{56.53} \\
Visual Aggregation Only & 50.27 & 30.12 & 86.61 & 55.67 \\
w/o Visual Aggregation & 40.80 & 30.40 & 78.08 & 49.76 \\
\mbox{w/o Cross-Invocation Selection} & 48.33 & 31.21 & 86.79 & 55.45 \\
w/o Joint Fitting & 47.93 & 30.81 & 86.76 & 55.17 \\
w/o Value Correction & 48.00 & 31.21 & 86.61 & 55.28 \\
\bottomrule
\end{tabularx}
\end{table}
\endgroup

\paragraph{Observation count and correction timing.}
We vary observation count to examine whether more call information offsets later selection, correction, and compaction.
Table~\ref{tab:ablation_observation_count} uses the first $m=1,\ldots,6$ shared-layer calls for selection and fitting on LoopVL-1B at $87.5\%$ target average pruning, fixing the initial pool and adjusting the final budget for later compaction.
Newly evaluated counts score $54.72$--$55.40\%$ Avg.\ (3) without monotonic gains; compression time rises from $31.10$ to $95.67$\,ms.
$m=2$ yields $56.53\%$ at $42.05$\,ms versus $55.36\%$ at $95.67$\,ms for $m=6$.
Two observations provide cross-call information while letting correction affect later computation sooner, explaining our choice (App.~C).

\begin{table}[!htbp]
\centering
\caption{\textbf{Observation-count ablation.} Scores (\%) and compression/prefill times ($T_c/T_p$, ms) with $m$ calls.}
\label{tab:ablation_observation_count}
\setlength{\tabcolsep}{1.5pt}
\renewcommand{\arraystretch}{.96}
\footnotesize
\begin{tabularx}{\linewidth}{@{}c*{6}{>{\centering\arraybackslash}X}@{}}
\toprule
$m$ & MMStar & \shortstack{MMMU\\-Pro} & POPE & \shortstack{Avg.\\(3)} & $T_c$ & $T_p$ \\
\midrule
1 & 48.27 & 31.27 & 86.65 & 55.40 & 31.10 & 209.66 \\
\rowcolor{softcyan}\textbf{2} & \textbf{51.00} & 31.62 & 86.98 & \textbf{56.53} & \textbf{42.05} & \textbf{219.43} \\
3 & 47.07 & 30.75 & 86.34 & 54.72 & 55.47 & 235.09 \\
4 & 47.07 & 31.27 & 86.23 & 54.85 & 68.64 & 251.16 \\
5 & 47.20 & \textbf{31.85} & 86.41 & 55.15 & 81.40 & 257.71 \\
6 & 47.20 & 31.79 & \textbf{87.10} & 55.36 & 95.67 & 272.66 \\
\bottomrule
\end{tabularx}
\end{table}

\paragraph{Hyperparameter Sensitivity.}
At $87.5\%$ average pruning, Figure~\ref{fig:hyperparameter_main} tests nine non-default settings of the instruction-query cap $P$, regularization $\lambda$, and candidate-pool direct-retention fraction $\alpha$, using MMStar/MMMU-Pro/POPE Avg.\ (3).
Varying one parameter with the others at default: $P\in\{2,4,16\}$ yields $55.29\%$--$55.64\%$, non-default $\lambda$ settings yield $55.05\%$--$55.23\%$, and $\alpha=0.70$ and $0.95$ yield $56.36\%$ and $55.49\%$, respectively (App.~I).
Among the reported configurations, the defaults ($P=8$, $\lambda=10^{-4}$, $\alpha=0.85$) achieve the highest Avg.\ (3) of $56.53\%$; we therefore adopt these settings.

\subsection{Further Analysis}
\label{sec:further_analysis}

After performance comparisons and component ablations, we revisit Section~\ref{sec:observations}: \textbf{why must token scoring and value correction consider different shared-layer calls?} Figure~\ref{fig:further_analysis} analyzes token coverage and output deviations in LoopVL-1B's first shared attention layer ($75\%$ local pruning), supporting cross-call selection and joint fitting.

We use $160$ questions: $40$ each from HRBench4K, ChartQA, MMStar, and POPE. Diagnostics fix $25\%$ visual tokens and all text tokens across calls, evaluating recorded uncompressed states without correction (App.~A).

\paragraph{Cross-invocation selection.}
In Figure~\ref{fig:further_analysis}(a--c), first-call selection covers $99$ of call $2$'s $222$ high-attention tokens ($44.6\%$), versus $169$ ($76.1\%$) for two-call selection on the MMStar example. High-attention windshield patches rise from $9$ to $23$ between these calls.
Across the cohort, two-call selection improves coverage at each call $3$--$6$ and reduces mean relative output deviation from $8.10\%$ to $4.82\%$, supporting cross-invocation selection (Section~\ref{sec:response_contribution}).

\paragraph{Call-dependent output deviations.}
With first-call selection fixed (Figure~\ref{fig:further_analysis}d), median maximum-to-minimum output-deviation ratios over calls $3$--$6$ are $4.2\times$, $2.5\times$, $3.4\times$, and $5.1\times$ on the four benchmarks (Figure~\ref{fig:further_analysis}e).
Identical removal affects call states differently, motivating joint fitting to both calls' original outputs (Section~\ref{sec:value_correction}).

\section{Conclusion}
\method combines visual aggregation, cross-invocation selection, and weight-tied value correction for training-free compression of looped VLMs.
Across two models and eleven benchmarks, it achieves leading average scores among compared compression methods with lower inference latency and memory usage.

\label{main:text_end}
\clearpage
\section*{LLM Usage Statement}
We used GPT and Codex for idea exploration, experimental planning, literature search, mathematical reasoning, writing and language refinement, and coding assistance.
The authors independently verified all AI-assisted content and take full responsibility for the methods, results, and conclusions.

{
\small

\clearpage
}

\clearpage
\appendix
\section*{Appendix}
\pdfbookmark[0]{Appendix}{appendix_overview}
\makeatletter
\raggedbottom
\setcounter{dbltopnumber}{2}
\setcounter{dblbotnumber}{1}
\setcounter{topnumber}{3}
\setcounter{bottomnumber}{1}
\setcounter{totalnumber}{5}
\renewcommand{\dbltopfraction}{0.86}
\renewcommand{\topfraction}{0.85}
\renewcommand{\bottomfraction}{0.50}
\renewcommand{\textfraction}{0.06}
\renewcommand{\floatpagefraction}{0.76}
\renewcommand{\dblfloatpagefraction}{0.76}
\setlength{\textfloatsep}{7pt plus 1pt minus 1pt}
\setlength{\dbltextfloatsep}{8pt plus 1pt minus 1pt}
\setlength{\floatsep}{9pt plus 2pt minus 2pt}
\setlength{\dblfloatsep}{8pt plus 1pt minus 1pt}
\setlength{\@fptop}{0pt}
\setlength{\@dblfptop}{0pt}
\setlength{\@fpsep}{12pt plus 2pt minus 2pt}
\setlength{\@dblfpsep}{12pt plus 2pt minus 2pt}
\setlength{\@fpbot}{0pt plus 1fil}
\setlength{\@dblfpbot}{0pt plus 1fil}
\makeatother

\section{Observations on 160 Questions}
\label{app:obs_scale}

\paragraph{Protocol.}
For Figure~\ref{fig:observation_detailed}(c--e), we sample $40$ questions each from HRBench4K, ChartQA, MMStar, and POPE (seed $20261101$). In recorded uncompressed states, we retain the same $25\%$ of visual tokens and all text tokens across six calls. We compare selection by first-call attention, mean attention from the first two calls, and random selection. Coverage is the fraction of each call's top-$25\%$ visual tokens retained. Output deviation is the relative Frobenius difference from that call's uncompressed output, with values unchanged.

\paragraph{Results.}
Two-call selection improves coverage at each of calls $3$--$6$ (Figure~\ref{fig:observation_detailed}(c)) and lowers mean output deviation from $8.10\%$ to $4.82\%$.

\begin{table}[H]
\centering\footnotesize
\caption{\textbf{Per-benchmark statistics at $75\%$ local pruning.} Results cover calls $3$--$6$. Spread is the median max/min error ratio for first-call selection.}
\label{tab:obs_scale}
\begin{tabular*}{\columnwidth}{@{\extracolsep{\fill}}lccccc@{}}
\toprule
& \multicolumn{2}{c}{Coverage (\%)} & & \multicolumn{2}{c}{Error (\%)} \\
\cmidrule(lr){2-3}\cmidrule(lr){5-6}
Benchmark & First & Random & Spread & First & First two \\
\midrule
HRBench4K & 38.1 & 24.9 & $4.2\times$ & 8.3 & 4.9 \\
ChartQA & 35.8 & 25.7 & $2.5\times$ & 9.6 & 5.4 \\
MMStar & 44.7 & 25.2 & $3.4\times$ & 5.8 & 4.1 \\
POPE & 37.6 & 25.1 & $5.1\times$ & 8.7 & 4.8 \\
\midrule
All & 39.0 & 25.2 & -- & 8.1 & 4.8 \\
\bottomrule
\end{tabular*}
\end{table}
\section{Train-Example Output Deviation}
\label{app:mechanism_diagnostic}
\label{app:observation_details}

The stars in Figure~\ref{fig:observation_detailed}(d) mark the open-ended train-number question from HRBench4K~\cite{wang2025hrbench} (image $240$). We use the first call's attention to retain $1980$ of $7921$ visual keys ($75\%$ local pruning) and keep the same set and all nonvisual keys, with values unchanged. At calls $3$--$6$ of the first shared attention layer, we measure the relative Frobenius deviation from the full-key output after gating and output projection.
\section{Additional Ablation Details}
\label{app:planned_ablation_protocol}

Table~\ref{tab:ablation_observation_count} compares the first $m=1,\ldots,6$ calls of the same shared attention layer on LoopVL-1B at $87.5\%$ target average pruning. We fix $K_0$ and use the default hyperparameters, adjusting $K_1$ for the compaction point. The observed calls guide token selection and joint fitting. We apply value correction during the last observed call and compact the sequence after its block.

Prefill time $T_p$ includes compression and excludes input preparation. Compression time $T_c$ excludes backbone computation between observations.
\section{Detailed Model Architectures}
\label{app:model_setup}
\label{app:baselines_and_settings}

\paragraph{LoopVL-1B.}
LoopVL-1B~\cite{qian2026loopvl} has two $16$-layer stacks. Each of the two cycles runs the low-level stack three times and the high-level stack once, totaling $128$ layer calls. LoopZip observes the low-level stack's first attention layer at zero-based calls $0$ and $16$.

\paragraph{SenseNova-Loop-9B-Preview.}
SenseNova-Loop-9B-Preview~\cite{sensenova2026loop} executes layers $4$--$27$ of its $32$-layer decoder twice, for $56$ layer calls. LoopZip observes the first full-attention layer in the loop (zero-based layer $7$) at calls $7$ and $31$. It corrects values at call $31$ and compacts both residual streams before call $32$.
\section{Detailed Benchmark Results}
\label{app:detailed_results}
\label{app:more_results}
\begin{table*}[tb]
 \centering
 \caption{\textbf{LoopVL-1B at $75\%$ and $87.5\%$ pruning ratios.} Scores are percentages. Rel.\ Avg.\ averages ratios to Vanilla. Bold marks the best results among compression methods at each pruning ratio.}
 \label{tab:main}
 \setlength{\tabcolsep}{1pt}
 \renewcommand{\arraystretch}{1.0}
 \footnotesize
 \begin{tabularx}{\textwidth}{@{}l*{13}{>{\centering\arraybackslash}X}@{}}
\toprule
 & \multicolumn{6}{c}{\textbf{Visual Understanding}} & \multicolumn{2}{c}{\textbf{Hallucination}} & \multicolumn{3}{c}{\textbf{Reasoning}} & \multicolumn{2}{c}{} \\
\cmidrule(lr){2-7}\cmidrule(lr){8-9}\cmidrule(lr){10-12}
\textbf{Method} & \textbf{\shortstack{MM\\Star}} & \textbf{RWQA} & \textbf{VMCB} & \textbf{AI2D} & \textbf{\shortstack{Chart\\QA}} & \textbf{\shortstack{HR\\Bench\\4K}} & \textbf{POPE} & \textbf{Hallu.} & \textbf{\shortstack{MMMU\\-Pro}} & \textbf{MathV} & \textbf{LogicV} & \textbf{Avg.$\uparrow$} & \textbf{\shortstack{Rel.\\Avg.$\uparrow$}} \\
\midrule
\rowcolor{softgray}Vanilla & 63.13 & 70.85 & 70.50 & 75.78 & 74.60 & 56.50 & 90.84 & 44.41 & 38.32 & 38.22 & 43.18 & 60.58 & 100.00\% \\
\midrule
\multicolumn{14}{c}{\cellcolor{softblue!45}\textbf{Pruning ratio $p=75\%$}} \\
\midrule
FastV~\cite{chen2024fastv}~(ECCV'24) & 44.93 & 49.28 & 56.40 & 63.12 & 24.68 & 43.25 & 75.28 & 35.22 & 33.58 & 31.05 & 37.14 & 44.90 & 75.52\% \\
SparseVLM~\cite{zhang2025sparsevlm}~(ICML'25) & 44.27 & 49.41 & 58.80 & 63.44 & 22.60 & 44.88 & 76.33 & 33.14 & 32.14 & 31.32 & 35.57 & 44.72 & 74.87\% \\
CDPruner~\cite{zhang2025cdpruner}~(NeurIPS'25) & 47.73 & 60.78 & 60.40 & 69.20 & 26.76 & 52.88 & 87.12 & 34.22 & 33.06 & 30.36 & 33.78 & 48.75 & 80.43\% \\
VisionZip~\cite{yang2025visionzip}~(CVPR'25) & 51.40 & 63.01 & 64.90 & 69.62 & 43.48 & 56.00 & 90.02 & 38.26 & 34.91 & 33.39 & 37.14 & 52.92 & 87.40\% \\
VisionTrim~\cite{yu2026visiontrim}~(ICLR'26) & 51.87 & 59.74 & 62.80 & 68.62 & 38.40 & 54.62 & 87.93 & 38.47 & \textbf{35.20} & 31.97 & 36.91 & 51.50 & 85.33\% \\
IVC-Prune~\cite{sun2026ivcprune}~(ICLR'26) & 46.20 & 53.86 & 56.60 & 62.66 & 22.80 & 44.62 & 83.31 & 32.51 & 31.62 & 30.43 & 33.11 & 45.25 & 75.04\% \\
V$^2$Drop~\cite{chen2026v2drop}~(CVPR'26) & 47.33 & 56.47 & 60.40 & 64.22 & 36.00 & 49.38 & 84.79 & 33.38 & 28.55 & 27.53 & 35.79 & 47.62 & 78.06\% \\
RESTORE~\cite{cho2026restore}~(ICML'26) & 53.67 & 63.27 & 62.70 & 70.01 & 49.84 & 54.62 & \textbf{90.24} & 38.72 & 34.22 & \textbf{34.21} & 34.23 & 53.25 & 87.61\% \\
QViD~\cite{athi2026vision}~(NeurIPS'26) & 48.67 & 57.91 & 65.70 & 65.51 & 51.12 & 55.88 & 86.31 & 38.72 & 33.29 & 32.60 & 34.90 & 51.87 & 85.55\% \\
\addlinespace[2pt]
\rowcolor{softcyan}\textbf{\method (Ours)} & \textbf{54.60} & \textbf{64.71} & \textbf{66.60} & \textbf{72.96} & \textbf{60.60} & \textbf{58.25} & 89.01 & \textbf{41.37} & 34.22 & 33.95 & \textbf{39.15} & \textbf{55.95} & \textbf{92.07\%} \\
\midrule
\multicolumn{14}{c}{\cellcolor{softblue!45}\textbf{Pruning ratio $p=87.5\%$}} \\
\midrule
FastV~\cite{chen2024fastv}~(ECCV'24) & 39.47 & 47.19 & 47.60 & 59.26 & 15.00 & 37.25 & 66.80 & 27.90 & 29.48 & 29.54 & 34.00 & 39.41 & 66.38\% \\
SparseVLM~\cite{zhang2025sparsevlm}~(ICML'25) & 38.33 & 44.84 & 50.10 & 59.16 & 13.44 & 36.88 & 70.04 & 29.16 & 28.79 & 29.84 & 35.57 & 39.65 & 66.80\% \\
CDPruner~\cite{zhang2025cdpruner}~(NeurIPS'25) & 43.47 & 55.29 & 56.50 & 64.31 & 17.80 & 54.25 & 85.18 & 31.75 & 27.17 & 27.63 & 31.32 & 44.97 & 73.89\% \\
VisionZip~\cite{yang2025visionzip}~(CVPR'25) & 47.00 & 58.95 & 60.90 & 64.60 & 27.16 & \textbf{55.50} & \textbf{88.13} & 34.76 & 31.16 & 30.10 & 35.12 & 48.49 & 80.06\% \\
VisionTrim~\cite{yu2026visiontrim}~(ICLR'26) & 47.73 & 55.56 & 59.80 & 63.92 & 24.48 & 49.38 & 83.72 & 34.88 & 30.00 & 29.61 & \textbf{36.02} & 46.83 & 77.57\% \\
IVC-Prune~\cite{sun2026ivcprune}~(ICLR'26) & 37.33 & 44.71 & 50.00 & 58.68 & 12.76 & 36.62 & 75.95 & 26.97 & 26.01 & 27.57 & 31.99 & 38.96 & 64.63\% \\
V$^2$Drop~\cite{chen2026v2drop}~(CVPR'26) & 41.00 & 47.32 & 52.00 & 59.78 & 20.68 & 41.75 & 79.27 & 25.73 & 21.16 & 21.51 & 31.77 & 40.18 & 65.11\% \\
RESTORE~\cite{cho2026restore}~(ICML'26) & 46.67 & 57.65 & 59.30 & 64.28 & 32.56 & 52.88 & 88.08 & 36.29 & 30.58 & 31.05 & 34.00 & 48.48 & 79.99\% \\
QViD~\cite{athi2026vision}~(NeurIPS'26) & 41.87 & 51.37 & 60.90 & 61.72 & 32.00 & 46.62 & 76.24 & 35.63 & 30.29 & \textbf{31.32} & 33.33 & 45.57 & 75.86\% \\
\addlinespace[2pt]
\rowcolor{softcyan}\textbf{\method (Ours)} & \textbf{51.00} & \textbf{59.74} & \textbf{64.90} & \textbf{66.61} & \textbf{44.12} & 55.12 & 86.98 & \textbf{37.96} & \textbf{31.62} & 31.02 & 32.89 & \textbf{51.09} & \textbf{83.89\%} \\
\bottomrule
\end{tabularx}
\end{table*}

\paragraph{Results at different pruning ratios.}
Table~\ref{tab:main} reports the per-benchmark results for Figure~\ref{fig:aggregate_results}. At $75\%$ pruning, \method achieves $55.95$ Avg.\ and $92.07\%$ Rel.\ Avg., exceeding RESTORE by $2.70$ and $4.46$ percentage points and ranking first on eight of eleven benchmarks. At $87.5\%$ pruning, \method achieves $51.09$ Avg.\ and $83.89\%$ Rel.\ Avg., exceeding VisionZip by $2.60$ and $3.83$ points and ranking first on seven benchmarks. These results show that \method preserves model performance at both higher pruning ratios.
\section{Efficiency Protocol and Additional Results}
\label{app:efficiency_details}
\begin{table*}[tb]
\centering
\caption{\textbf{Efficiency at $N=2048$ visual tokens.} P, D, E, and M denote prefill (ms), decode (ms/token), end-to-end latency (s), and peak allocated GPU memory (GiB), respectively. Bold highlights LoopZip.}
\label{tab:efficiency_n2048}
\begingroup
\scriptsize
\setlength{\tabcolsep}{2.5pt}
\renewcommand{\arraystretch}{1.12}
\begin{tabular*}{\textwidth}{@{\extracolsep{\fill}}l*{12}{r}@{}}
\toprule
Method & \multicolumn{4}{c}{HF} & \multicolumn{4}{c}{vLLM} & \multicolumn{4}{c}{SGLang} \\
\cmidrule(lr){2-5}\cmidrule(lr){6-9}\cmidrule(l){10-13}
 & P & D & E & M & P & D & E & M & P & D & E & M \\
\midrule
Vanilla & 374 & 17.0 & 0.92 & 4.73 & 293 & 10.3 & 0.64 & 4.79 & 271 & 10.1 & 0.61 & 4.80 \\
\midrule
\multicolumn{13}{l}{Average pruning ratio: $50\%$} \\
VisionZip & 260 & 16.1 & 0.78 & 3.97 & 193 & 8.6 & 0.49 & 3.99 & 197 & 8.7 & 0.50 & 4.00 \\
VisionTrim & 224 & 18.1 & 0.80 & 3.97 & 197 & 9.5 & 0.52 & 3.99 & 202 & 9.8 & 0.53 & 4.00 \\
FastV & 325 & 17.9 & 0.90 & 3.97 & 193 & 9.4 & 0.51 & 4.03 & 197 & 9.4 & 0.52 & 4.04 \\
SparseVLM & 324 & 18.0 & 0.90 & 3.97 & 220 & 8.6 & 0.51 & 4.23 & 231 & 9.5 & 0.55 & 4.24 \\
CDPruner & 638 & 18.0 & 1.21 & 3.97 & 423 & 8.8 & 0.72 & 3.99 & 450 & 9.8 & 0.78 & 4.00 \\
IVC-Prune & 353 & 18.0 & 0.93 & 3.96 & 198 & 8.6 & 0.49 & 4.03 & 202 & 9.4 & 0.52 & 4.04 \\
V$^2$Drop & 361 & 17.9 & 0.94 & 3.97 & 198 & 9.4 & 0.52 & 4.04 & 203 & 10.2 & 0.55 & 4.04 \\
RESTORE & 255 & 26.5 & 1.10 & 3.97 & 200 & 18.7 & 0.81 & 3.99 & 211 & 19.2 & 0.84 & 4.00 \\
QViD & 328 & 18.0 & 0.90 & 3.97 & 215 & 8.6 & 0.51 & 4.03 & 221 & 9.7 & 0.55 & 4.04 \\
\textbf{\method} & \textbf{323} & \textbf{17.9} & \textbf{0.90} & \textbf{3.95} & \textbf{208} & \textbf{9.5} & \textbf{0.53} & \textbf{4.00} & \textbf{213} & \textbf{9.1} & \textbf{0.52} & \textbf{4.01} \\
\midrule
\multicolumn{13}{l}{Average pruning ratio: $75\%$} \\
VisionZip & 186 & 15.5 & 0.69 & 3.56 & 174 & 9.0 & 0.48 & 3.59 & 187 & 8.7 & 0.49 & 3.59 \\
VisionTrim & 224 & 17.3 & 0.78 & 3.56 & 176 & 9.0 & 0.48 & 3.59 & 183 & 9.8 & 0.51 & 3.59 \\
FastV & 189 & 15.5 & 0.69 & 3.58 & 172 & 9.0 & 0.48 & 3.66 & 176 & 8.7 & 0.47 & 3.66 \\
SparseVLM & 222 & 17.3 & 0.78 & 3.58 & 187 & 9.0 & 0.49 & 3.86 & 194 & 8.7 & 0.49 & 3.87 \\
CDPruner & 317 & 17.3 & 0.87 & 3.56 & 273 & 9.0 & 0.58 & 3.59 & 288 & 9.7 & 0.62 & 3.60 \\
IVC-Prune & 201 & 17.2 & 0.75 & 3.56 & 176 & 9.0 & 0.48 & 3.66 & 188 & 8.7 & 0.49 & 3.66 \\
V$^2$Drop & 203 & 17.2 & 0.76 & 3.59 & 175 & 9.0 & 0.48 & 3.66 & 181 & 9.6 & 0.51 & 3.67 \\
RESTORE & 187 & 22.1 & 0.89 & 3.54 & 175 & 14.3 & 0.65 & 3.63 & 190 & 15.0 & 0.69 & 3.63 \\
QViD & 212 & 17.3 & 0.77 & 3.58 & 192 & 9.0 & 0.50 & 3.66 & 197 & 8.7 & 0.50 & 3.66 \\
\textbf{\method} & \textbf{202} & \textbf{17.2} & \textbf{0.76} & \textbf{3.55} & \textbf{187} & \textbf{9.0} & \textbf{0.49} & \textbf{3.60} & \textbf{197} & \textbf{9.8} & \textbf{0.53} & \textbf{3.60} \\
\midrule
\multicolumn{13}{l}{Average pruning ratio: $87.5\%$} \\
VisionZip & 195 & 15.5 & 0.69 & 3.38 & 164 & 8.8 & 0.46 & 3.38 & 170 & 9.7 & 0.50 & 3.39 \\
VisionTrim & 258 & 15.4 & 0.75 & 3.38 & 167 & 8.8 & 0.47 & 3.38 & 175 & 9.7 & 0.51 & 3.39 \\
FastV & 192 & 15.3 & 0.69 & 3.36 & 166 & 8.8 & 0.47 & 3.47 & 176 & 8.7 & 0.47 & 3.48 \\
SparseVLM & 225 & 17.1 & 0.78 & 3.38 & 177 & 8.0 & 0.45 & 3.68 & 189 & 9.8 & 0.52 & 3.68 \\
CDPruner & 253 & 17.0 & 0.80 & 3.38 & 211 & 8.8 & 0.51 & 3.40 & 221 & 9.7 & 0.55 & 3.41 \\
IVC-Prune & 229 & 17.0 & 0.78 & 3.33 & 172 & 8.5 & 0.46 & 3.47 & 182 & 9.7 & 0.51 & 3.47 \\
V$^2$Drop & 222 & 17.1 & 0.77 & 3.36 & 168 & 8.0 & 0.44 & 3.48 & 174 & 8.8 & 0.48 & 3.48 \\
RESTORE & 216 & 19.9 & 0.85 & 3.38 & 172 & 12.2 & 0.58 & 3.44 & 188 & 12.7 & 0.61 & 3.45 \\
QViD & 227 & 17.1 & 0.78 & 3.36 & 192 & 8.8 & 0.49 & 3.48 & 194 & 9.8 & 0.53 & 3.48 \\
\textbf{\method} & \textbf{221} & \textbf{17.1} & \textbf{0.77} & \textbf{3.39} & \textbf{177} & \textbf{8.8} & \textbf{0.48} & \textbf{3.39} & \textbf{186} & \textbf{9.8} & \textbf{0.52} & \textbf{3.40} \\
\bottomrule
\end{tabular*}
\endgroup
\end{table*}
\begin{table*}[tb]
\centering
\caption{\textbf{Efficiency at $N=4096$ visual tokens.} P, D, E, and M denote prefill (ms), decode (ms/token), end-to-end latency (s), and peak allocated GPU memory (GiB), respectively. Bold highlights LoopZip.}
\label{tab:efficiency_n4096}
\begingroup
\scriptsize
\setlength{\tabcolsep}{2.5pt}
\renewcommand{\arraystretch}{1.12}
\begin{tabular*}{\textwidth}{@{\extracolsep{\fill}}l*{12}{r}@{}}
\toprule
Method & \multicolumn{4}{c}{HF} & \multicolumn{4}{c}{vLLM} & \multicolumn{4}{c}{SGLang} \\
\cmidrule(lr){2-5}\cmidrule(lr){6-9}\cmidrule(l){10-13}
 & P & D & E & M & P & D & E & M & P & D & E & M \\
\midrule
Vanilla & 729 & 18.2 & 1.32 & 6.36 & 633 & 11.9 & 1.05 & 6.41 & 557 & 10.5 & 0.93 & 6.42 \\
\midrule
\multicolumn{13}{l}{Average pruning ratio: $50\%$} \\
VisionZip & 501 & 18.8 & 1.11 & 4.76 & 359 & 10.3 & 0.72 & 4.81 & 334 & 10.1 & 0.69 & 4.82 \\
VisionTrim & 433 & 19.0 & 1.05 & 4.76 & 364 & 10.3 & 0.73 & 4.81 & 340 & 10.1 & 0.70 & 4.82 \\
FastV & 371 & 17.1 & 0.93 & 4.79 & 354 & 10.3 & 0.72 & 4.89 & 332 & 9.4 & 0.67 & 4.90 \\
SparseVLM & 469 & 19.0 & 1.09 & 4.79 & 410 & 10.3 & 0.77 & 5.65 & 388 & 9.1 & 0.71 & 5.65 \\
CDPruner & 1040 & 18.9 & 1.66 & 4.76 & 635 & 10.2 & 1.00 & 4.83 & 642 & 10.1 & 1.00 & 4.84 \\
IVC-Prune & 459 & 18.8 & 1.08 & 4.76 & 361 & 10.3 & 0.72 & 4.89 & 337 & 10.1 & 0.69 & 4.90 \\
V$^2$Drop & 545 & 18.9 & 1.16 & 4.78 & 355 & 10.2 & 0.72 & 4.90 & 331 & 10.3 & 0.69 & 4.91 \\
RESTORE & 511 & 35.1 & 1.63 & 4.77 & 402 & 27.5 & 1.30 & 4.89 & 400 & 28.1 & 1.32 & 4.89 \\
QViD & 466 & 19.0 & 1.09 & 4.79 & 380 & 10.3 & 0.74 & 4.89 & 356 & 10.3 & 0.72 & 4.90 \\
\textbf{\method} & \textbf{480} & \textbf{19.2} & \textbf{1.11} & \textbf{4.73} & \textbf{377} & \textbf{10.3} & \textbf{0.74} & \textbf{4.82} & \textbf{362} & \textbf{10.2} & \textbf{0.72} & \textbf{4.83} \\
\midrule
\multicolumn{13}{l}{Average pruning ratio: $75\%$} \\
VisionZip & 280 & 16.1 & 0.81 & 3.98 & 253 & 9.4 & 0.59 & 4.00 & 256 & 9.8 & 0.60 & 4.01 \\
VisionTrim & 294 & 16.5 & 0.84 & 3.98 & 258 & 9.5 & 0.60 & 4.00 & 263 & 9.8 & 0.61 & 4.01 \\
FastV & 267 & 17.8 & 0.85 & 4.01 & 252 & 9.4 & 0.59 & 4.14 & 255 & 8.8 & 0.57 & 4.15 \\
SparseVLM & 328 & 17.7 & 0.91 & 4.07 & 285 & 8.6 & 0.60 & 4.90 & 291 & 8.8 & 0.61 & 4.90 \\
CDPruner & 678 & 16.0 & 1.21 & 3.98 & 495 & 9.4 & 0.84 & 4.08 & 504 & 9.8 & 0.86 & 4.09 \\
IVC-Prune & 391 & 16.4 & 0.93 & 3.98 & 256 & 9.4 & 0.59 & 4.14 & 260 & 9.9 & 0.61 & 4.14 \\
V$^2$Drop & 369 & 17.9 & 0.96 & 4.02 & 258 & 9.3 & 0.59 & 4.15 & 261 & 10.0 & 0.62 & 4.16 \\
RESTORE & 351 & 24.6 & 1.15 & 3.98 & 264 & 18.7 & 0.89 & 4.19 & 274 & 19.3 & 0.92 & 4.19 \\
QViD & 351 & 17.8 & 0.93 & 4.01 & 276 & 9.4 & 0.61 & 4.14 & 281 & 9.9 & 0.63 & 4.14 \\
\textbf{\method} & \textbf{356} & \textbf{18.1} & \textbf{0.95} & \textbf{3.95} & \textbf{270} & \textbf{9.5} & \textbf{0.61} & \textbf{4.00} & \textbf{277} & \textbf{9.9} & \textbf{0.63} & \textbf{4.01} \\
\midrule
\multicolumn{13}{l}{Average pruning ratio: $87.5\%$} \\
VisionZip & 249 & 15.4 & 0.76 & 3.57 & 232 & 9.0 & 0.56 & 3.59 & 237 & 9.7 & 0.58 & 3.60 \\
VisionTrim & 256 & 15.5 & 0.77 & 3.57 & 236 & 9.0 & 0.56 & 3.60 & 244 & 9.8 & 0.59 & 3.61 \\
FastV & 243 & 15.5 & 0.75 & 3.63 & 231 & 9.0 & 0.56 & 3.77 & 239 & 8.9 & 0.56 & 3.77 \\
SparseVLM & 337 & 15.5 & 0.85 & 4.07 & 250 & 9.0 & 0.57 & 4.53 & 256 & 9.9 & 0.61 & 4.53 \\
CDPruner & 371 & 15.3 & 0.88 & 3.57 & 331 & 9.0 & 0.66 & 3.70 & 355 & 9.7 & 0.71 & 3.71 \\
IVC-Prune & 302 & 15.3 & 0.81 & 3.59 & 235 & 9.0 & 0.56 & 3.77 & 240 & 9.7 & 0.59 & 3.77 \\
V$^2$Drop & 275 & 15.5 & 0.79 & 3.65 & 234 & 9.0 & 0.56 & 3.78 & 239 & 9.9 & 0.59 & 3.78 \\
RESTORE & 278 & 20.3 & 0.94 & 3.56 & 240 & 14.1 & 0.73 & 3.84 & 252 & 14.9 & 0.76 & 3.84 \\
QViD & 290 & 15.4 & 0.80 & 3.63 & 255 & 9.0 & 0.58 & 3.77 & 260 & 9.8 & 0.61 & 3.77 \\
\textbf{\method} & \textbf{277} & \textbf{17.5} & \textbf{0.85} & \textbf{3.55} & \textbf{244} & \textbf{9.0} & \textbf{0.57} & \textbf{3.60} & \textbf{253} & \textbf{9.8} & \textbf{0.60} & \textbf{3.60} \\
\bottomrule
\end{tabular*}
\endgroup
\end{table*}
\begin{table*}[tb]
\centering
\caption{\textbf{Efficiency at $N=8192$ visual tokens.} P, D, E, and M denote prefill (ms), decode (ms/token), end-to-end latency (s), and peak allocated GPU memory (GiB), respectively. Bold highlights LoopZip.}
\label{tab:efficiency_n8192}
\begingroup
\scriptsize
\setlength{\tabcolsep}{2.5pt}
\renewcommand{\arraystretch}{1.12}
\begin{tabular*}{\textwidth}{@{\extracolsep{\fill}}l*{12}{r}@{}}
\toprule
Method & \multicolumn{4}{c}{HF} & \multicolumn{4}{c}{vLLM} & \multicolumn{4}{c}{SGLang} \\
\cmidrule(lr){2-5}\cmidrule(lr){6-9}\cmidrule(l){10-13}
 & P & D & E & M & P & D & E & M & P & D & E & M \\
\midrule
Vanilla & 1581 & 20.7 & 2.28 & 9.61 & 1539 & 15.0 & 2.08 & 9.65 & 1316 & 11.6 & 1.76 & 9.66 \\
\midrule
\multicolumn{13}{l}{Average pruning ratio: $50\%$} \\
VisionZip & 919 & 18.4 & 1.55 & 6.42 & 849 & 11.8 & 1.30 & 6.45 & 771 & 10.5 & 1.18 & 6.46 \\
VisionTrim & 948 & 18.4 & 1.57 & 6.42 & 857 & 11.4 & 1.29 & 6.45 & 780 & 10.6 & 1.19 & 6.46 \\
FastV & 878 & 18.5 & 1.51 & 6.46 & 822 & 11.8 & 1.27 & 6.60 & 746 & 9.4 & 1.12 & 6.61 \\
SparseVLM & 982 & 18.4 & 1.61 & 6.59 & 941 & 11.9 & 1.39 & 9.62 & 873 & 10.6 & 1.28 & 9.63 \\
CDPruner & 1257 & 18.3 & 1.89 & 6.42 & 1151 & 11.8 & 1.60 & 6.74 & 1094 & 10.5 & 1.50 & 6.75 \\
IVC-Prune & 1043 & 18.2 & 1.67 & 6.41 & 839 & 11.8 & 1.29 & 6.60 & 763 & 10.5 & 1.17 & 6.61 \\
V$^2$Drop & 882 & 18.4 & 1.51 & 6.46 & 810 & 11.8 & 1.26 & 6.63 & 744 & 10.8 & 1.16 & 6.64 \\
RESTORE & 1058 & 50.7 & 2.69 & 6.45 & 981 & 44.3 & 2.44 & 6.92 & 980 & 45.6 & 2.47 & 6.92 \\
QViD & 924 & 18.6 & 1.56 & 6.46 & 857 & 11.9 & 1.30 & 6.60 & 779 & 10.6 & 1.19 & 6.61 \\
\textbf{\method} & \textbf{964} & \textbf{20.7} & \textbf{1.66} & \textbf{6.34} & \textbf{874} & \textbf{11.8} & \textbf{1.33} & \textbf{6.47} & \textbf{800} & \textbf{10.6} & \textbf{1.21} & \textbf{6.47} \\
\midrule
\multicolumn{13}{l}{Average pruning ratio: $75\%$} \\
VisionZip & 614 & 17.2 & 1.20 & 4.80 & 566 & 9.8 & 0.96 & 4.82 & 540 & 10.1 & 0.93 & 4.83 \\
VisionTrim & 669 & 17.2 & 1.26 & 4.80 & 576 & 10.3 & 0.98 & 4.82 & 549 & 10.1 & 0.94 & 4.83 \\
FastV & 554 & 17.1 & 1.14 & 4.88 & 543 & 10.0 & 0.93 & 5.10 & 516 & 10.1 & 0.91 & 5.11 \\
SparseVLM & 659 & 17.2 & 1.25 & 6.59 & 630 & 10.2 & 1.04 & 8.12 & 600 & 10.1 & 0.99 & 8.13 \\
CDPruner & 992 & 17.0 & 1.58 & 4.80 & 864 & 10.1 & 1.26 & 5.24 & 861 & 10.1 & 1.25 & 5.25 \\
IVC-Prune & 772 & 17.0 & 1.36 & 4.81 & 557 & 10.2 & 0.95 & 5.10 & 526 & 10.1 & 0.92 & 5.11 \\
V$^2$Drop & 687 & 18.9 & 1.34 & 4.87 & 551 & 10.3 & 0.95 & 5.13 & 525 & 10.3 & 0.93 & 5.13 \\
RESTORE & 687 & 35.3 & 1.84 & 4.81 & 617 & 27.6 & 1.56 & 5.62 & 616 & 27.0 & 1.54 & 5.63 \\
QViD & 664 & 18.8 & 1.30 & 4.88 & 574 & 10.2 & 0.97 & 5.10 & 548 & 10.2 & 0.95 & 5.11 \\
\textbf{\method} & \textbf{788} & \textbf{19.1} & \textbf{1.44} & \textbf{4.73} & \textbf{580} & \textbf{10.3} & \textbf{0.98} & \textbf{4.83} & \textbf{557} & \textbf{10.2} & \textbf{0.96} & \textbf{4.83} \\
\midrule
\multicolumn{13}{l}{Average pruning ratio: $87.5\%$} \\
VisionZip & 482 & 16.2 & 1.04 & 4.00 & 472 & 9.5 & 0.85 & 4.01 & 463 & 9.8 & 0.85 & 4.02 \\
VisionTrim & 479 & 16.1 & 1.03 & 4.00 & 468 & 9.4 & 0.84 & 4.04 & 472 & 9.9 & 0.86 & 4.05 \\
FastV & 468 & 16.1 & 1.03 & 4.10 & 452 & 9.4 & 0.83 & 4.35 & 447 & 9.8 & 0.83 & 4.36 \\
SparseVLM & 537 & 16.2 & 1.10 & 6.59 & 493 & 9.3 & 0.87 & 7.37 & 497 & 9.9 & 0.88 & 7.38 \\
CDPruner & 707 & 16.0 & 1.27 & 4.00 & 671 & 8.6 & 1.02 & 4.49 & 690 & 9.8 & 1.07 & 4.50 \\
IVC-Prune & 511 & 16.0 & 1.06 & 4.02 & 453 & 9.3 & 0.82 & 4.35 & 455 & 9.9 & 0.85 & 4.36 \\
V$^2$Drop & 567 & 16.2 & 1.14 & 4.09 & 461 & 9.5 & 0.84 & 4.38 & 460 & 10.1 & 0.86 & 4.38 \\
RESTORE & 528 & 24.6 & 1.37 & 4.16 & 483 & 18.0 & 1.12 & 4.98 & 494 & 18.3 & 1.14 & 4.99 \\
QViD & 574 & 16.2 & 1.15 & 4.10 & 476 & 9.3 & 0.85 & 4.35 & 476 & 9.9 & 0.87 & 4.36 \\
\textbf{\method} & \textbf{520} & \textbf{17.9} & \textbf{1.13} & \textbf{3.94} & \textbf{476} & \textbf{9.3} & \textbf{0.85} & \textbf{4.02} & \textbf{476} & \textbf{9.9} & \textbf{0.87} & \textbf{4.02} \\
\bottomrule
\end{tabular*}
\endgroup
\end{table*}

\paragraph{Measurement setup.}
After two warmups, we time five fixed MMStar inputs at $2048$, $4096$, or $8192$ visual tokens, generating $32$ tokens greedily. All tests use NVIDIA H20 96GB GPUs, batch size $1$, and BF16 with FP32 adapters. Each method uses the same GPU across backends, and Vanilla and LoopZip are measured on the same GPU.

\paragraph{Backends and output checks.}
In Hugging Face Transformers~\cite{wolf2020transformers}, prefill uses PyTorch~\cite{paszke2019pytorch} FlashAttention~\cite{dao2022flashattention}. vLLM~\cite{kwon2023vllm} uses FlashAttention-2~\cite{dao2024flashattention2} prefill and Triton paged decoding. SGLang~\cite{zheng2024sglang} uses FlashInfer~\cite{ye2025flashinfer} through RadixAttention. The serving decoders replay a full CUDA graph for each single-token decode step, using fused operators and unquantized weights. On the timing records, vLLM and SGLang's answer choices agree with HF on $451/464$ and $453/464$ records, respectively.

\paragraph{Metrics.}
Prefill timing starts when the batch is ready. Decode latency averages $31$ cached steps, end-to-end latency runs from image loading to text output, and GPU memory averages peak CUDA allocation per input. Archived LoopZip timings include the original error checks. Tables~\ref{tab:efficiency_n2048}--\ref{tab:efficiency_n8192} report the full comparisons for Figure~\ref{fig:efficiency}.
\section{Visual-Token Pruning Baselines}
\label{app:baseline_implementation}

\paragraph{Matched budgets.}
Within each model, all methods use the same prompts, image preprocessing, and generation limits (Appendices~\ref{app:model_setup} and~\ref{app:evaluation}). Pruning preserves all text tokens. For a baseline that prunes once after $d$ full-length calls, we set
\begin{equation}
  K_d=\operatorname{clip}_{[1,N]}\!\left(
  \left\lfloor N\frac{L(1-p)-d}{L-d}+\frac12\right\rfloor\right),
\end{equation}
where $N$ is the original visual length, $p$ is the target average pruning ratio, and $L$ is the total layer-call count ($128$ for LoopVL-1B and $56$ for SenseNova). For multistage pruning, we match the same budget by weighting each retained count by the number of calls that use it.

\paragraph{Pruning schedules.}
On LoopVL-1B, FastV~\cite{chen2024fastv}, SparseVLM~\cite{zhang2025sparsevlm}, IVC-Prune~\cite{sun2026ivcprune}, and QViD~\cite{athi2026vision} prune before the third layer call. CDPruner~\cite{zhang2025cdpruner} prunes before the backbone. On SenseNova, these four methods read the first full-attention layer and prune before zero-based call $4$. IVC-Prune also compacts earlier caches, with earlier linear-attention calls counted at full length on SenseNova. In both models, V$^2$Drop~\cite{chen2026v2drop} prunes after $2$, $8$, and $11$ layer calls.
\section{Evaluation Settings and Metrics}
\label{app:evaluation}

\paragraph{Benchmark sizes and generation.}
Table~\ref{tab:app_data} gives the input counts and output-token limits for the eleven benchmarks in Section~\ref{sec:experimental_setup}. All methods within each model share prompts, preprocessing, and generation limits.

\paragraph{Evaluation settings.}
SenseNova runs with thinking disabled. On ChartQA and POPE it uses a minimum image area of $1{,}048{,}576$ pixels, with short numeric or label answers for ChartQA and yes/no answers for POPE. These settings were selected on training data and shared by all methods.

\begin{table}[H]
  \centering
  \caption{\textbf{Benchmark sizes and generation ceilings.} For LoopVL-1B, columns headed $8$, $32$, and $512$ count inputs assigned each maximum output-token budget. SenseNova-Loop-9B-Preview uses $1024$ for every input.}
  \label{tab:app_data}
  \footnotesize
  \setlength{\tabcolsep}{3pt}
  \begin{tabular}{lrrrr}
    \toprule
    Benchmark & Inputs & 8 & 32 & 512 \\
    \midrule
    \multicolumn{5}{c}{\cellcolor{softblue!45}\textit{Visual understanding}} \\
    MMStar & 1500 & 0 & 1500 & 0 \\
    RealWorldQA & 765 & 0 & 765 & 0 \\
    VMCBench (DEV) & 1000 & 0 & 1000 & 0 \\
    AI2D (TEST) & 3088 & 0 & 3087 & 1 \\
    ChartQA (TEST) & 2500 & 0 & 2500 & 0 \\
    HRBench4K & 800 & 0 & 800 & 0 \\
    \midrule
    \multicolumn{5}{c}{\cellcolor{softblue!45}\textit{Hallucination}} \\
    POPE & 5127 & 0 & 5127 & 0 \\
    HallusionBench & 951 & 560 & 385 & 6 \\
    \midrule
    \multicolumn{5}{c}{\cellcolor{softblue!45}\textit{Reasoning}} \\
    MMMU-Pro & 1730 & 0 & 1729 & 1 \\
    MathVision & 3040 & 0 & 3040 & 0 \\
    LogicVista & 447 & 0 & 412 & 35 \\
    \midrule
    Total & 20{,}948 & 560 & 20{,}345 & 43 \\
    \bottomrule
  \end{tabular}
\end{table}

\paragraph{Scoring.}
We use a fixed VLMEvalKit-based matcher~\cite{duan2024vlmevalkit}. HRBench4K averages scores across four option orders per question, with unparsed answers counted as incorrect. ChartQA uses a $0.05$ numerical tolerance. POPE pools F1 over $9000$ instances, and HallusionBench averages aAcc, fAcc, and qAcc.

\paragraph{Aggregate scores.}
For each benchmark $j$, let $s_j$ and $s_j^{\mathrm{vanilla}}$ be the percentage scores of a compression method and Vanilla.
\begin{equation}
  \begin{aligned}
    \mathrm{Avg.}&=\frac{1}{11}\sum_{j=1}^{11}s_j,\\
    \mathrm{Rel.\ Avg.}&=\frac{100}{11}
    \sum_{j=1}^{11}\frac{s_j}{s_j^{\mathrm{vanilla}}}.
  \end{aligned}
\end{equation}

\section{Hyperparameter Sensitivity}
\label{app:hyperparameter_sensitivity}

At $87.5\%$ pruning, we vary one of $(P,\lambda,\alpha)=(8,10^{-4},0.85)$ at a time. Tables~\ref{tab:hyperparameter_queries}--\ref{tab:hyperparameter_alpha} detail the nine settings in Figure~\ref{fig:hyperparameter_main}. Avg.\ (3) averages MMStar and MMMU-Pro accuracy and POPE F1.

All ten settings are timed on the same $20$ MMStar inputs (seed $20261104$), with two warmups and three interleaved runs per setting. We report the mean paired prefill change, excluding input construction.

\begin{table}[H]
\centering
\caption{Instruction-query cap $P$; fixed $\lambda=10^{-4}$ and $\alpha=0.85$. Bold marks the default. Its latency was measured in this sweep. Scores are percentages; $\Delta t$ is paired prefill change relative to the timed default.}
\label{tab:hyperparameter_queries}
\label{tab:hyperparameter_sensitivity}
\setlength{\tabcolsep}{2pt}
\footnotesize
\begin{tabular}{@{}r|rrrr|rr@{}}
\toprule
$P$ & MMStar & MMMU-P & POPE & Avg.\ (3) & \shortstack{Prefill\\(ms)} & $\Delta t$ \\
\midrule
2 & 48.47\% & 31.85\% & 86.61\% & 55.64\% & 196.63 & +0.53\% \\
4 & 47.33\% & 31.68\% & 86.86\% & 55.29\% & 195.70 & +0.05\% \\
\textbf{8} & \textbf{51.00\%} & \textbf{31.62\%} & \textbf{86.98\%} & \textbf{56.53\%} & \textbf{195.59} & \textbf{+0.00\%} \\
16 & 48.13\% & 31.21\% & 87.07\% & 55.47\% & 195.63 & +0.00\% \\
\bottomrule
\end{tabular}
\end{table}

\begin{table}[H]
\centering
\caption{Ridge coefficient $\lambda$; fixed $P=8$ and $\alpha=0.85$. Bold marks the default. Its latency was measured in this sweep. Scores are percentages; $\Delta t$ is paired prefill change relative to the timed default.}
\label{tab:hyperparameter_ridge}
\setlength{\tabcolsep}{2pt}
\footnotesize
\begin{tabular}{@{}r|rrrr|rr@{}}
\toprule
$\lambda$ & MMStar & MMMU-P & POPE & Avg.\ (3) & \shortstack{Prefill\\(ms)} & $\Delta t$ \\
\midrule
$10^{-6}$ & 47.60\% & 30.92\% & 86.65\% & 55.06\% & 194.97 & -0.31\% \\
$10^{-5}$ & 47.87\% & 30.69\% & 86.59\% & 55.05\% & 196.38 & +0.40\% \\
$\mathbf{10^{-4}}$ & \textbf{51.00\%} & \textbf{31.62\%} & \textbf{86.98\%} & \textbf{56.53\%} & \textbf{195.59} & \textbf{+0.00\%} \\
$10^{-3}$ & 47.87\% & 31.21\% & 86.60\% & 55.23\% & 195.96 & +0.19\% \\
$10^{-2}$ & 47.67\% & 30.98\% & 86.64\% & 55.10\% & 190.17 & -2.80\% \\
\bottomrule
\end{tabular}
\end{table}

\begin{table}[H]
\centering
\caption{Frontend dominant fraction $\alpha$; fixed $P=8$ and $\lambda=10^{-4}$. Bold marks the default. Its latency was measured in this sweep. Scores are percentages; $\Delta t$ is paired prefill change relative to the timed default.}
\label{tab:hyperparameter_alpha}
\setlength{\tabcolsep}{2pt}
\footnotesize
\begin{tabular}{@{}r|rrrr|rr@{}}
\toprule
$\alpha$ & MMStar & MMMU-P & POPE & Avg.\ (3) & \shortstack{Prefill\\(ms)} & $\Delta t$ \\
\midrule
0.70 & 49.93\% & 32.02\% & 87.14\% & 56.36\% & 195.08 & -0.27\% \\
\textbf{0.85} & \textbf{51.00\%} & \textbf{31.62\%} & \textbf{86.98\%} & \textbf{56.53\%} & \textbf{195.59} & \textbf{+0.00\%} \\
0.95 & 49.47\% & 30.17\% & 86.84\% & 55.49\% & 194.91 & -0.34\% \\
\bottomrule
\end{tabular}
\end{table}
\begin{figure*}[!t]
\centering
\includegraphics[width=\textwidth,trim=4bp 0bp 4bp 0bp,clip]{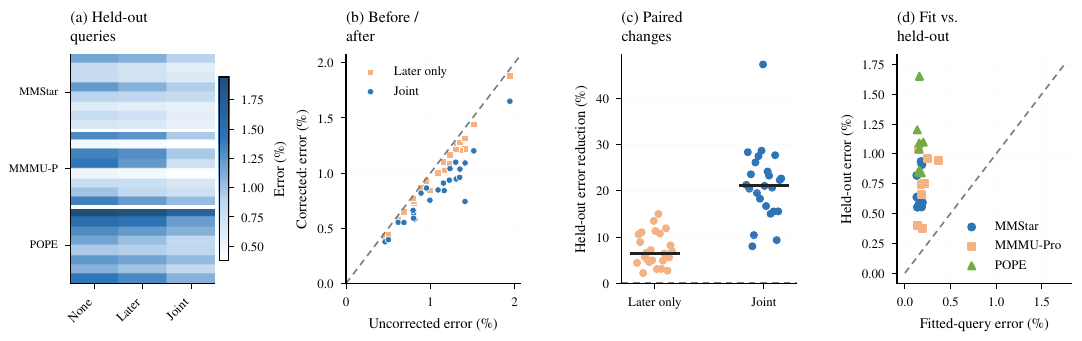}
\caption{\textbf{Value-correction transfer with disjoint instruction queries.} (a) Held-out relative output error for every input and all three variants; (b) paired before/after errors, with the diagonal denoting unchanged error; (c) per-input relative error reduction, including any negative changes; (d) joint-fit errors on fitting versus held-out queries. Both domains average calls $0$ and $16$. All variants share the same retained keys; held-out queries are excluded from fitting and acceptance checks.}
\label{fig:appendix_heldout_queries}
\end{figure*}

\section{Properties and Implementation}
\label{app:theoretical_analysis}
\label{app:method_details}

These details supplement the aggregation, fitting, and cache update in Section~\ref{sec:method}.

\subsection{Visual Candidate Pool Settings}
\label{app:representative_details}
For Section~\ref{sec:recurrent_compression}, round $\alpha K_0$ to the nearest integer and clip it to $[1,K_0]$ to obtain the number of directly retained patches. If slots remain, assign one center to each. Sample the centers from the remaining patches in raster order with stride equal to the number of remaining patches divided by the number of centers, rounded down.

\subsection{Two-Stage Token Budget}
\label{app:budget_details}
To implement Equation~\eqref{eq:method_average_budget}, first round the target average count to $K=\min(N,\max(1,\lfloor(1-p)N+1/2\rfloor))$. Then set
\begin{equation}
 \begin{aligned}
  \delta&=\min\{16,\max(1,\lfloor K/16\rfloor),N-K\},\\
  K_0&=K+\delta,\qquad
  K_1=\left\lfloor\frac{LK-dK_0}{L-d}\right\rfloor.
 \end{aligned}
 \label{eq:method_budget}
\end{equation}
If $\delta=0$ or $K_1\notin[1,K_0-1]$, set $K_0=K_1=K$. The realized visual-query pruning ratio is $p_Q=1-[dK_0+(L-d)K_1]/(LN)$. For LoopVL-1B, $17$ layer calls process $K_0$ visual queries and $111$ process $K_1$. The cache instead stores $16$ slots with $K_0$ visual keys and values and $112$ with $K_1$, because the second observed call writes compact KV while still processing all of its query rows. Table~\ref{tab:app_budget_examples} gives an example for each model.

\begin{table}[H]
\centering\footnotesize
\setlength{\tabcolsep}{3pt}
\caption{\textbf{Token budgets for an illustrative $N=2048$ input.} $p$ is the target, and $p_Q$ is the realized visual-query pruning ratio. Both are percentages.}
\label{tab:app_budget_examples}
\begin{tabular}{lrrrrrr}
\toprule
Model & $p$ & $L$ & $d$ & $K_0$ & $K_1$ & $p_Q$ \\
\midrule
LoopVL-1B & 87.5 & 128 & 17 & 272 & 253 & 87.52 \\
SenseNova-9B & 50 & 56 & 32 & 1040 & 1002 & 50.01 \\
\bottomrule
\end{tabular}
\end{table}

\subsection{Native Response Fitting}
\label{app:response_coordinates}
\label{app:value_fitting}
For Equation~\eqref{eq:method_response_fit}, remove the discarded visual key columns and renormalize attention over the retained keys using the original mask. With the queries, keys, and native gates fixed, the corrected response is
\begin{equation}
 \widehat{\mathbf O}_t(\mathbf X)
 =\widehat{\mathbf O}_t(0)+\mathcal L_t(\mathbf X).
 \label{eq:method_response_map}
\end{equation}
Here $\widehat{\mathbf O}_t(0)$ is the deletion-only response. The map $\mathcal L_t$ applies the retained visual attention weights, the native gate, and the output projection. A value-head correction is shared across both calls. In SenseNova, each correction also serves all query heads in its KV group.

Stack the matrix forms of the two maps and the vectorized residuals $\mathbf O_t-\widehat{\mathbf O}_t(0)$. Scale both stacks by $1/\sqrt{2P}$ to obtain $\mathbf B$ and $\boldsymbol b$. With $\boldsymbol x=\operatorname{vec}(\mathbf X)$, the normal equations are
\begin{equation}
 (\mathbf B^\top\mathbf B+\lambda\mathbf I)\boldsymbol x^\star
 =\mathbf B^\top\boldsymbol b .
 \label{eq:appendix_normal_equation}
\end{equation}
We solve them in FP32 by conjugate gradients from zero, with at most $32$ steps and a relative residual tolerance of $10^{-5}$.

\subsection{Installing the Value Correction}
\label{sec:response_analysis}
At $t=1$, we use the sampled instruction rows to compute the full-pool fitting reference. We then select the retained keys and jointly fit $\mathbf X^\star$. A single scale $\gamma\in[0,1]$ minimizes the second-call fitting error before casting, with $\gamma=0$ when the fitted direction leaves the response unchanged.

We add $\gamma\mathbf X^\star$ to the retained visual values, cast them to the native cache dtype, and store them with the retained keys and the text KV in the second call's cache slot. The second call completes attention over the retained keys using all original query rows. After the block, the hidden sequence is shortened for subsequent calls, as in Section~\ref{sec:loopzip_algorithm}. The first call's output and cache are preserved.

\section{More Analysis}
\label{app:more_analysis}
\subsection{Value Correction on Disjoint Queries}
\label{app:heldout_queries}

We sample eight inputs each from MMStar, MMMU-Pro, and POPE ($24$ total, seed $20261104$). On the recorded traces, we fix the keys selected by calls $(0,16)$ and hold out eight instruction queries disjoint from the eight fit queries. We compare no correction, later-only fitting, and joint fitting with matched normalization, ridge strength, and solver settings. Historical BF16 acceptance checks use only fitting queries.

Figure~\ref{fig:appendix_heldout_queries} shows held-out errors of $1.07\%$, $1.00\%$, and $0.84\%$, respectively. The joint fit reduces held-out error on all $24$ inputs. This experiment measures transfer within frozen responses.
\subsection{Value Correction in the Forward Pass}
\label{app:live_value_correction}

We reuse the $24$ LoopVL-1B inputs from Appendix~\ref{app:heldout_queries} at $87.5\%$ target pruning and fix the archived $K_0$ and $K_1$ positions. All variants share one candidate representation per input and compact at the same point. Against a reference that keeps $K_0$, we measure relative Frobenius error on eight held-out instruction rows of the native attention output at calls $16$ and $32$. Relative to no correction, joint fitting lowers mean error by $0.115$ and $0.028$ percentage points, respectively.

\begin{table}[H]
\centering\footnotesize
\caption{Mean relative output error (\%) in actual forward runs, paired over $24$ inputs. Bold marks the lowest mean per call.}
\label{tab:live_value_correction}
\begin{tabular}{lrr}
\toprule
Value correction & Second call & Next call \\
\midrule
None & 0.905 & 1.272 \\
Later-only fit & 0.805 & 1.270 \\
Joint fit & \textbf{0.790} & \textbf{1.244} \\
\bottomrule
\end{tabular}
\end{table}
\FloatBarrier
\section{SenseNova at Higher Pruning Ratios}
\label{app:sense_higher_pruning}

Table~\ref{tab:sense_higher_pruning} compares methods on six benchmarks covering visual understanding, hallucination, and reasoning. At $75\%$ average pruning, \method achieves $66.83$ Avg.\ and $89.93\%$ Rel.\ Avg., exceeding the strongest baseline by $2.32$ and $3.11$ percentage points, respectively. It ranks first on three benchmarks and ties for first on LogicVista. At $87.5\%$, the two averages are $58.81$ and $79.25\%$, exceeding the strongest baseline by $1.76$ and $2.37$ percentage points, with three first-place scores.

\begin{table*}[tb]
 \centering

 \caption{\textbf{Results at $75\%$ and $87.5\%$ pruning ratios.} Scores are percentages. Rel.\ Avg.\ averages ratios to Vanilla. Bold and underlined values indicate the best and second-best results among compression methods, respectively.}
 \label{tab:sense_higher_pruning}
 \setlength{\tabcolsep}{1pt}
 \renewcommand{\arraystretch}{0.88}
 \footnotesize
\begin{tabularx}{\textwidth}{@{}l*{8}{>{\centering\arraybackslash}X}@{}}
\toprule
 & \multicolumn{3}{c}{\textbf{Visual Understanding}} & \multicolumn{2}{c}{\textbf{Hallucination}} & \multicolumn{1}{c}{\textbf{Reasoning}} & \multicolumn{2}{c}{} \\
\cmidrule(lr){2-4}\cmidrule(lr){5-6}\cmidrule(lr){7-7}
\textbf{Method} & \textbf{RWQA} & \textbf{ChartQA} & \textbf{HRBench4K} & \textbf{POPE} & \textbf{Hallu.} & \textbf{LogicV} & \textbf{Avg.$\uparrow$} & \textbf{Rel. Avg.$\uparrow$} \\
\midrule
\rowcolor{softgray}Vanilla & 72.42 & 85.16 & 77.00 & 87.56 & 58.28 & 61.74 & 73.69 & 100.00\% \\
\midrule
\multicolumn{9}{c}{\cellcolor{softblue!45}\textbf{Pruning ratio $p=75\%$}} \\
\midrule
FastV~\cite{chen2024fastv}~(ECCV'24) & 58.69 & 13.92 & 75.75 & 75.66 & 41.06 & 36.91 & 50.33 & 68.74\% \\
SparseVLM~\cite{zhang2025sparsevlm}~(ICML'25) & \underline{67.97} & 50.00 & \textbf{78.12} & \underline{87.42} & 49.55 & 37.81 & 61.81 & 83.36\% \\
CDPruner~\cite{zhang2025cdpruner}~(NeurIPS'25) & 59.08 & 44.80 & 74.88 & 84.52 & 47.55 & 37.36 & 58.03 & 78.34\% \\
VisionZip~\cite{yang2025visionzip}~(CVPR'25) & 57.12 & 29.84 & 64.00 & 81.11 & 40.90 & \textbf{42.28} & 52.54 & 71.39\% \\
VisionTrim~\cite{yu2026visiontrim}~(ICLR'26) & 59.08 & 38.60 & 72.25 & 81.27 & 41.29 & 34.45 & 54.49 & 73.37\% \\
IVC-Prune~\cite{sun2026ivcprune}~(ICLR'26) & 64.44 & 28.64 & 71.25 & 84.00 & 45.32 & 38.93 & 55.43 & 75.31\% \\
V$^2$Drop~\cite{chen2026v2drop}~(CVPR'26) & 62.35 & 52.60 & 72.75 & 85.45 & 48.63 & 38.26 & 60.01 & 80.89\% \\
RESTORE~\cite{cho2026restore}~(ICML'26) & 66.80 & \underline{65.40} & 74.50 & \textbf{87.89} & \underline{51.97} & \underline{40.49} & \underline{64.51} & \underline{86.82\%} \\
QViD~\cite{athi2026vision}~(NeurIPS'26) & 64.58 & 30.44 & \underline{77.12} & 84.64 & 43.89 & 33.11 & 55.63 & 75.11\% \\
\addlinespace[2pt]
\rowcolor{softcyan}\textbf{\method (Ours)} & \textbf{69.28} & \textbf{74.40} & 76.25 & 85.53 & \textbf{53.24} & \textbf{42.28} & \textbf{66.83} & \textbf{89.93\%} \\
\midrule
\multicolumn{9}{c}{\cellcolor{softblue!45}\textbf{Pruning ratio $p=87.5\%$}} \\
\midrule
FastV~\cite{chen2024fastv}~(ECCV'24) & 45.23 & 8.52 & 66.50 & 51.93 & 31.50 & 37.36 & 40.17 & 55.45\% \\
SparseVLM~\cite{zhang2025sparsevlm}~(ICML'25) & \textbf{63.92} & 22.24 & \textbf{74.88} & \underline{84.40} & 41.86 & \underline{38.26} & 54.26 & 73.63\% \\
CDPruner~\cite{zhang2025cdpruner}~(NeurIPS'25) & 54.38 & 28.08 & 67.38 & 78.99 & 38.75 & 34.90 & 50.41 & 68.13\% \\
VisionZip~\cite{yang2025visionzip}~(CVPR'25) & 51.37 & 19.80 & 49.38 & 71.30 & 34.39 & 35.35 & 43.60 & 59.33\% \\
VisionTrim~\cite{yu2026visiontrim}~(ICLR'26) & 48.50 & 16.76 & 66.38 & 60.01 & 34.22 & 33.78 & 43.27 & 59.14\% \\
IVC-Prune~\cite{sun2026ivcprune}~(ICLR'26) & 53.73 & 15.96 & 60.38 & 73.93 & 38.87 & 34.00 & 46.14 & 62.92\% \\
V$^2$Drop~\cite{chen2026v2drop}~(CVPR'26) & 53.73 & 24.16 & 61.75 & 73.90 & 41.49 & 37.58 & 48.77 & 66.54\% \\
RESTORE~\cite{cho2026restore}~(ICML'26) & 62.09 & \underline{39.56} & 73.25 & \textbf{86.52} & \underline{43.51} & 37.36 & \underline{57.05} & \underline{76.88\%} \\
QViD~\cite{athi2026vision}~(NeurIPS'26) & 54.25 & 14.36 & 73.25 & 66.28 & 35.79 & 36.24 & 46.69 & 63.78\% \\
\addlinespace[2pt]
\rowcolor{softcyan}\textbf{\method (Ours)} & \underline{63.79} & \textbf{51.36} & \underline{73.62} & 80.76 & \textbf{44.85} & \textbf{38.48} & \textbf{58.81} & \textbf{79.25\%} \\
\bottomrule
\end{tabularx}
\end{table*}
\FloatBarrier
\begin{figure*}[!tb]
\section{Visual Token Selection Examples}
\label{app:visual_token_selection_gallery}

These examples show that the first two calls focus on different image regions and that averaging their attention can retain patches missed by either call alone. At a fixed $25\%$ retention budget, Figure~\ref{fig:visual_token_selection_gallery} compares the patches selected using each call's attention and their mean. For each benchmark, we use the first archived input whose original patch grid matches the recorded token count.
\par\smallskip
\centering
\includegraphics[width=\textwidth]{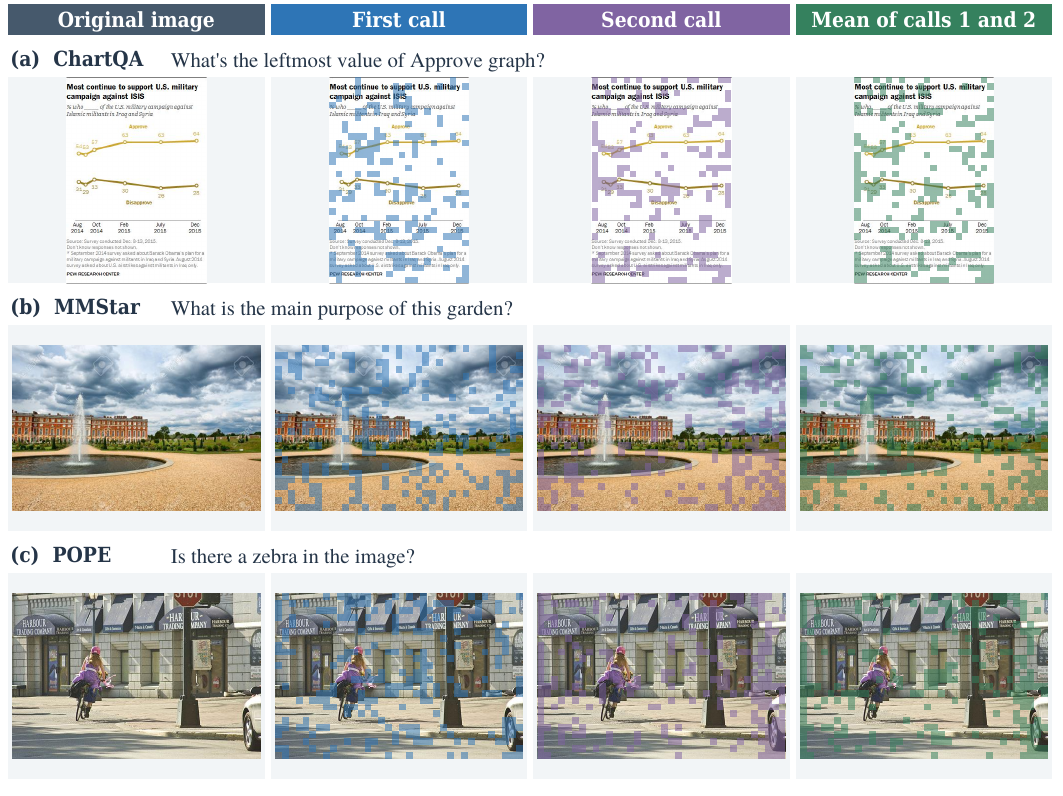}
\caption{\textbf{Selected visual patches on ChartQA, MMStar, and POPE.} Colored cells mark patches selected by each scoring rule. The masks use the original patch grid before visual aggregation.}
\label{fig:visual_token_selection_gallery}
\end{figure*}

\end{document}